\documentclass{cup-pan}
\usepackage[utf8]{inputenc}
\usepackage{blindtext}
\usepackage{marvosym}
\usepackage{ifsym}
\usepackage{graphicx}
\usepackage{microtype}
\usepackage{graphicx}
\usepackage{booktabs} 

\usepackage{adjustbox}
\usepackage{float}
\usepackage[T1,T2A]{fontenc}

\usepackage{dcolumn}
\usepackage{mathtools}
\usepackage{amssymb}
\usepackage{siunitx}
\usepackage{hyperref}
\usepackage[utf8]{inputenc}
\usepackage{csquotes}
\usepackage{comment}
\usepackage{color}
\usepackage{soul}
\usepackage{blindtext}
\usepackage{parskip}
\usepackage{adjustbox}
\usepackage{float}

\usepackage{graphicx}
\usepackage{booktabs} 

\usepackage{adjustbox}
\usepackage{float}

\usepackage{mathtools}
\usepackage{color}
\usepackage{soul}
\usepackage{amsmath}
\usepackage{amssymb}
\usepackage{siunitx}
\usepackage{hyperref}
\usepackage{cleveref}
\usepackage{csquotes}
\usepackage{comment}
\usepackage{hyperref}
\usepackage{mathtools}
\usepackage{color}
\usepackage{soul}
\usepackage{amsmath}
\usepackage{amssymb}
\usepackage{siunitx}
\usepackage{hyperref}
\usepackage{cleveref}
\usepackage{csquotes}
\usepackage{comment}

\title{Multi-aspect Multilingual and Cross-lingual Parliamentary Speech Analysis}

\author[1]{Kristian Miok}
\author[2]{Encarnación Hidalgo-Tenorio}
\author[3]{Petya Osenova}
\author[4]{Miguel-Ángel Benítez-Castro}
\author[5]{Marko Robnik-\v{S}ikonja}

\affil[1]{Computer Science Department, West University
of Timisoara, Bulevardul Vasile Pârvan 4,
300223 Timisoara, Romania. Email: \url{kristian.miok@e-uvt.ro}}
\affil[2]{Department of English and German Studies, Facultad de Filosofía y Letras, Universidad de Granada, Calle Prof. Clavera s/n, 18071, Granada, Spain. Email: \url{ehidalgo@ugr.es}}
\affil[3]{Faculty of Slavic Studies, Sofia University St. Kl. Ohridski, Sofia, Bulgaria.}
\affil[3]{Institute of Information and Communication Technologies, Bulgarian Academy of Sciences. Email: \url{petya@bultreebank.org}}
\affil[4]{Department of English and German Studies, Faculty of Social Sciences and Humanities, University of Zaragoza, Calle Ciudad Escolar, S/N, 44003, Teruel, Spain. Email: \url{mbenitez@unizar.es}}
\affil[5]{Faculty of Computer and Information Science, University of Ljubljana, Večna pot 113, 1000 Ljubljana, Slovenia. Email: \url{marko.robnik@fri.uni-lj.si}}

\corrauthor{Kristian Miok}

\runningauthor{Miok et al.}

\begin{document}

\maketitle

\begin{abstract}

Parliamentary and legislative debate transcripts provide informative insight into elected politicians' opinions, positions, and policy preferences. They are interesting for political and social sciences as well as linguistics and natural language processing (NLP) research. While existing research studied individual parliaments, we apply advanced NLP methods to a joint and comparative analysis of six national parliaments (Bulgarian, Czech, French, Slovene, Spanish, and United Kingdom) between 2017 and 2020. We analyze emotions and sentiment in the transcripts from the ParlaMint dataset collection and assess if the age, gender, and political orientation of speakers can be detected from their speeches. The results show some commonalities and many surprising differences among the analyzed countries.

\end{abstract}
\section{Introduction}
\label{intro}

The development of natural language resources and technologies opens new possibilities for social and political sciences \cite{rheault2016measuring}. After decades of analyzing individual political speeches and transcripts, natural language processing (NLP) allows orders of magnitude larger studies. Parliamentary corpora are available for many parliamentary democracies and include draft bills, amendments to bills, adopted legislation, committee reports, and transcripts of floor debates. Processing these heterogeneous records is challenging. However, the recent ParlaMint project has produced unified corpora of parliamentary debates in 17 European parliaments, making them widely accessible \cite{erjavec2022parlamint}. This allows for broadening the scope of analyses from individual countries to joint issues and differences. Modern monolingual and cross-lingual NLP techniques applied to the collected data can provide new insight into the language used, the expression of speakers, as well as similarities and differences in emotions and sentiment in different parliaments. 

National parliaments have developed their own code of behaviour and speech \citet{maurer2001national}. The comparison between them is difficult. We propose a novel technological approach, combining monolingual and cross-lingual prediction models with machine translation to analyze political, sociological, and linguistic phenomena.  Recent research has shown that such approaches are possible for analysis of social media but they have not yet been applied to parliamentary speech corpora. We analyze parliamentary debates from six national parliaments: Bulgarian, Czech, English, French, Slovene, Spanish, and United Kingdom (UK) in the period from 2017 to 2020.


The main contributions of our work can be summarized as follows:
\begin{enumerate}
    \item A methodological framework for a comprehensive comparison of parliamentary speeches in a cross-lingual setting using cross-domain transfer learning.
    \item Comparison of linguistic effects based on age, gender, and political position of speakers for six parliaments.
    \item Comparison of sentiment and emotions in six parliaments extracted with prediction models.
    \item A novel approach to analyse differences in sentiment based on speakers' age using Bayesian statistical models. 
\end{enumerate}


The paper is structured into five further sections. In Section \ref{sec:relatedWork}, we present background and related work on analyzing parliamentary speech split into age, gender, sentiment and emotions.  
Section \ref{sec:data} describes the used ParlaMint datasets. In Section \ref{sec:methodology}, we present our methodology, split into a prediction of metadata (age, gender, and political wing), sentiment, and emotions. The results are covered in Section \ref{sec:results}, while we draw the conclusions and present ideas for further work in Section \ref{sec:conclusions}.

\section{Background and Related Work}
\label{sec:relatedWork}
In this section, we first present the necessary background information, discussing also aspects not strictly related to the parliamentary context. Sections \ref{related:age}, \ref{related:gender}, \ref{related:sentiment}, and \ref{related:emotions} cover related works referring to age, gender, sentiment, and emotions in the political discourse. Finally, in Section \ref{relatedBayesian}, we outline the use of Bayesian methods in parliamentary contexts.


\subsection{Language and Age}
\label{related:age}
Age as a factor of language variation is one of the most salient and productive objects of research in the field of sociolinguistics \cite{murphy2010corpus}. The description of differences between young and older generations focuses on (in)formality \cite{labov1972sociolinguistic,stenstrom2009youngspeak}. 
In the research of group membership through speech \citet{ghafournia2015language}, sociolinguists describe two types of prestige, overt and covert prestige. Overt prestige is related to standard and more formal linguistic features, which are normally associated with those who hold more power and status. Covert prestige, on the other hand, is the non-standard variety employed in a scenario that encourages cooperation, communality, communication ease, and engagement \cite{trudgill1972sex}. Based upon these considerations, some linguistic differences age may explain are adults' preference for syntactic complexity \cite{frizelle2018growth}, swearing \cite{jay2013child}, lexical conservativism \cite{kerswill1996children}, usage of positive politeness strategies \cite{emara2017gender}, teenagers' tendency towards language change \cite{milroy1985linguistic}, the use of slang \cite{rodriguez1994youth} or abruptness \cite{de2012youth}.

Several authors covered the prediction of age and other personal traits such as gender or political affiliation, e.g., \citet{dahllof2012automatic}, who analyzed the wording of political speeches in Swedish. The results show that it is possible to classify politicians according to their age, ideology, and gender to some degree. We analyze six parliaments at once, which opens a broader perspective and gives more general conclusions.

\subsection{Language and Gender}
\label{related:gender}
Since 1922, a number of studies have addressed the role of gender in language expression -- for an overview, see \citet{tenorio2016genderlect}. 
For example, the debate ranges on whether gender is a social construct, whether there exist different genderlects with different characteristics, and whether a so-called “women’s language” is the result of culture or power relations \cite{coates2015women},  \cite{lakoff1973language}. The linguistic features claimed to characterize females range from articulatory phonetics and grammar to pure pragmatics, e.g., the tendency for hypercorrection, conservativism, self-disclosure and attentiveness; abundance of intensifiers and restricted vocabulary associated with domesticity; preference for simple syntax, minimal responses, emotion(al) language, expressive speech acts, diminutives and terms of endearment; usage of rising intonation, questions and epistemic modality to mark their lack of confidence; and, finally, neither swearing nor turn-taking control, interruption or topic selection in conversation.

In our work, we predict the gender of speakers available as metadata. In this way, we establish a level of differences between speeches used by members of parliament (MPs) of different genders. In gender detection, we find some interesting research that successfully applies machine learning and/or sentiment analysis \cite{argamon2003gender,park2019gender,menendez2020damegender,kowsari2020gender}. An important consideration in the prediction of speakers’ gender is grammatical gender. In the four of the six languages we cover, Bulgarian, Czech, Slovenian, and Spanish, there are three grammatical genders (masculine, feminine, and neuter); in French, there are two genders (masculine and feminine), while in English has no grammatical gender. Grammatical gender can generally be inferred from the ending of nouns, adjectives, determiners or past participles. In some cases (not all), this means that the gender of a speaker can be determined.
Next, we give some examples of these phenomena for the analyzed languages.

\begin{description}
\item \textbf{BG}: Az sam sigurna (I am sure: feminine), Az sam siguren (I am sure: masculine). In Bulgarian, there
are synthetic and analytic tenses/moods. The former contains no indication of gender. Here is an
example of a synthetic form: Az kazax (I said - no gender marking, thus it might apply to either feminine or masculine). However, when an analytic form with a participle is used, the gender is indicated by this participle: Az bix predlozhila (I would suggest: feminine) vs. Az bih predlozhil (I would suggest: masculine).
\item \textbf{CZ}: J\'{a} bych \v{r}ekl (I would say: masculine), J\'{a} bych \v{r}ekla (I would say: feminine). Again, if a synthetic form is used, then there is no indication of a specific gender: J\'{a} si mysl\'{i}m (I think: feminine and masculine).
\item \textbf{ES}: Estoy harta del populismo (I'm sick of populism: feminine); Estoy harto del populismo (I'm sick of populism: masculine).
\item \textbf{FR}: Je suis prête (I am ready: feminine),  Je suis prêt (I am ready: masculine). Again, there are many cases where the gender is not revealed, e.g., Comme j’ai dit (As I said: feminine and masculine).
\item \textbf{SI}: The gender is revealed when using the first person singular in the past and future tense, e.g., Rekla sem (I said: feminine), Rekel sem (I said: masculine). The gender is not revealed in the present tense, e.g., Mislim (I think: both masculine and feminine).
\end{description}
 
 In English, gender is not a grammatical category but a lexico-sematic feature that can be inferred from the personal and relative pronouns used (the person who arrived; he is nice) and a few morphemes (actor vs actress; policeman vs policewoman); adjectives, determiners or past participles do not show it (this happy man vs this happy woman; she was kissed vs he was kissed). These features are not revealing of the speakers’ gender.

\subsection{Sentiment in Politics}
\label{related:sentiment}

The problem of computational sentiment analysis for parliament discourses has been tackled extensively but with relatively little cross-country comparison. In most cases, sentiment analysis involves document, sentence, and aspect-level analysis. 

The research of \citet{dziecikatko2018application} and \citet{rheault2016measuring} have applied sentiment analysis to entire corpora at the highest granularity, i.e. their analysis of the Polish and UK parliaments aggregate sentiment scores of all speeches. \citet{honkela2014five} explore the overall sentiment of EU Parliament transcripts on the dataset level, whereas \citet{sakamoto2017cross} consider the polarity of US and Japanese datasets. 

While in NLP sentiment analysis is often fine-grained (such as at the level of speech, speech segment, paragraph, sentence, or phrase), in political science, the unit of analysis is primarily an actor (individual politician whose contributions are pooled together). This is the focus of most works on position scaling, a task very much associated with that field. It appears that this confirms to some extent \citet{hopkins2010method} assertion that while computer scientists are interested in finding the needle in the haystack, social scientists are more interested in characterizing it. The exceptions come from works in the social and political sciences \cite{iliev2019political,hopkins2010method} that propose ways to optimize speech-level classification for social science purposes and from computer science \cite{glavavs2017unsupervised}, which also consider the position scaling issue.

Sentiment detection has advanced considerably in the last few years with the advent of large pre-trained language models such as BERT \cite{devlin-etal-2019-bert}. This has allowed applications to social media, stock market predictions, user stance detection in reviews, hate-speech detection, etc. However, parliamentary discourse is hard to analyze for established techniques due to specific formal speech and linguistic differences to existing training datasets \cite{rheault2016measuring}. \citet{rudkowskysupervised} study several machine learning approaches based on word embeddings for Austrian parliamentary speeches. Similarly, \citet{abercrombie2020parlvote} and \citet{elkink2021predicting} investigate predicting votes based on the parliament speeches.

In this paper, we follow political sciences and predict the sentiments of speakers based on their speeches. The results are cross-lingual for six different parliaments, which, to our knowledge, has not been done before.

\subsection{Emotions in Politics}
\label{related:emotions}
\citet{alba2018emotion} states that whatever we say, write, hear, and read is produced and processed through the filter of affect. Cognition and emotion are, therefore, two mutually interconnected systems \cite{barrett2020seven}.
In this regard, \citet{van1985handbook} argues that one of the most distinguishing features of manipulation lies in shaping and framing messages in such a way that they accord with their recipients’ negative emotions, usually deriving from feelings of powerlessness and injustice. 
In the current political landscape, which is imbued with populism, this idea is of utmost importance, especially at a moment when the emotional is preferred to the intellectual. 

Research shows how resentment, anxiety, panic, anger, and disgust can help populist politicians seduce their voters \cite{betz1993new}. 
They may use the same discursive strategies to attack and bring their rivals into disrepute. For instance, they can spread unreliable news about their opponents and other sensationalist information with bombastic but simple expressions; plentiful negative ethical and aesthetic evaluative terms; swear words and colloquialisms; and adversarial vocabulary echoing 20th-century propaganda\footnote{\url{https://www.thebritishacademy.ac.uk/blog/how-language-fake-news-echoes-20th-century-propaganda/}}.  Despite the similarities, however, the discourses of right- and left-wing populist leaders are quite diverse. Open opposition to capitalist elites drives left-wing populists to show their hatred of big corporations, financial, and governmental institutions \cite{de1997populism}. On the other hand, due to their fear of losing their status because of the alleged privileges granted to minority groups in a multicultural society, right-wing populists cannot conceal their antagonism and hostility towards such communities \cite{salmela2017emotional}. 

Our analysis is unique in detecting and comparing emotions in six national parliaments at once. This reveals some similarities but also surprising differences. 

\subsection{Bayesian Methods for Parliamentary Data}
\label{relatedBayesian}
Bayesian methods allow for the estimation of the posterior probability of a model-given data, which provides a measure of both the quality of the model and the uncertainty of the estimates. This is particularly useful when dealing with small data sets or noisy data.
In our previous work, we applied Monte Carlo Dropout to deal with prediction uncertainty \cite{miok2019prediction, miok2022ban} in natural language processing. Within the parliamentary context, \citet{hansen2009positions} estimated the positions of the Irish parliamentary parties using the Bayesian ideal point estimation framework and showed that it is possible to distinguish only two blocs of parties in each period, one block supporting the government and one forming the opposition. \citet{han2007analysing} explored the performance of spatial voting models to the roll calls of the EU parliament. \citet{montalvo2019bayesian} proposed a new methodology for predicting electoral results that combines a fundamental model and national polls within an evidence synthesis framework.   

In this work, we propose to use the Bayesian AB testing to determine the age threshold with the largest difference between positive and negative sentiment. In this way, we avoid setting a fixed threshold for all parliaments and get interesting differences between the tested parliaments.

\section{The Data}
\label{sec:data}
In this section, we describe the datasets used in our analysis. Section \ref{sec:parlamint} describes the ParlaMint project, which collected and preprocessed the data, while in Section \ref{sec:datasets} we provide information on the actually used datasets and graphically present distributions of MP's age and gender across parliaments.

\subsection{ParlaMint Project Background}
\label{sec:parlamint}

ParlaMint\footnote{\url{http://www.clarin.eu/parlamint}} project aims to enhance the development and usage of national parliamentary corpora. The data has been synchronized with respect to the same TEI format and time span. It can be exploited for linguistic, social, and political research in cross-lingual and cross-parliament settings.

We use the multilingual comparable corpora of parliamentary debates ParlaMint 2.1 containing parliamentary debates mostly starting in 2015 and extending to mid-2020, with each national corpus containing various amounts of words varying from \num{800000} words (for Hungarian) to \num{109000000} words (for UK). The sessions in the corpora are marked as belonging to the COVID-19 period (after November 1st 2019), or being "reference" (before that date).
The data is freely available through the CLARIN.SI repository\footnote{\url{http://hdl.handle.net/11356/1432}}.

The corpora contain extensive metadata, including many aspects of the speakers (name, gender, MP status, party affiliation, party coalition/opposition). The data are structured into time-stamped terms, sessions, and meetings. Speeches are marked by the speakers and their roles (e.g., the chair or regular speaker). The speeches also contain marked-up transcriber comments, such as gaps in the transcription, interruptions, applause, etc. More information about the creation of the corpora, the common standard, and the specifics of each national corpus can be found in \citet{erjavec2022parlamint}.

\subsection{The Datasets}
\label{sec:datasets}

At the time of writing this paper, the ParlaMint project released data for 16 languages: Bulgarian, Croatian, Czech, Danish, Dutch, English, French, Hungarian, Icelandic, Italian, Latvian, Lithuanian, Polish, Slovenian, Spanish, and Turkish. In total there are \num{3774204} utterances and \num{494949904} words. The quality of the textual corpora and metadata varies across the languages. In our experiments, we studied parliaments in six countries: Bulgaria (BG), Czech Republic (CZ), France (FR), Slovenia (SI), Spain (ES), and the United Kingdom (UK). The criteria for this selection were mainly the quality of the provided corpora and that we, as authors, understand the languages and the political situation in these countries. The available data for the specific year varies across the parliaments, as shown in Table \ref{tab:numberWords}. We decided to analyze data from 2017 to 2020 expecting this selection to be the most informative and provide the most interesting insights. 

The data per parliament and year are organized as parliament session documents. Every text document with the talks is paired with the document containing the session metadata such as title, time, term, number of sessions and meetings. This supplement document also includes the speaker information (speaker type, speaker party, party parliament status, speaker name, speaker gender, and speaker birth). The number of parliament session documents per country and year are presented in Table \ref{tab:numberSessions}.
The number of sessions varies per parliament. For the Czech parliament, the number is the largest, while the Slovene and Spanish parliament exhibit lower numbers of sessions.
\begin{table}[H]
	\caption{Number of words per parliament and per year in the analyzed part of the ParlaMint corpus.} 
	\centering
	\resizebox{\textwidth}{!}{
	\begin{tabular}{l|cccccc|c}
	 \textbf{Year} & \textbf{BG} & \textbf{CZ}& \textbf{ES}& \textbf{FR} &    \textbf{SI}  &  \textbf{UK} & $\Sigma$ \\
		\hline			
\textbf{2017}      &  3,052,523  & 2,959,956  & 2,553,736  & 5,584,411   & 4,489,429 &   16,715,363 & 35,355,418\\
\textbf{2018}      & 4,342,676   & 3,487,291  & 3,175,000  & 13,470,543  & 2,842,075 &  19,675,067 & 46,992,652\\
\textbf{2019}      & 3,349,615   & 4,003,544  & 885,134    & 11,864,705  & 3,601,000 &  16,937,349   & 40,641,347\\
\textbf{2020}      & 2,240,914   & 4,299,581  & 3,697,803  & 5,767,864   & 1,885,329 &   19,435,104 & 37,326,595\\ 
\hline
\textbf{$\Sigma$}  & 12,985,728  & 14,750,372 &  10,311,673  & 36,687,523  & 12,817,833  & 72,762,8834  & 160,316,012 \\ 
	\end{tabular}
}	
	\label{tab:numberWords}
\end{table}

\begin{table}[H]
	\caption{Number of the sessions per parliament and per year in the analyzed part of the ParliaMint corpus.} 
	\centering
	\renewcommand{\arraystretch}{1.2}
	\setlength{\tabcolsep}{6pt}
	\begin{tabular}{l|cccccc|c}
	 \textbf{Year} &  \textbf{BG} &  \textbf{CZ} & \textbf{ES} &  \textbf{FR} &    \textbf{SI}  &  \textbf{UK} & $\Sigma$ \\
		\hline			
\textbf{2017} & 101 & 711 &  64 & 156   & 74  & 259 & 1365\\
\textbf{2018} & 132 & 635 &  71 & 362   & 65 & 309 & 1574 \\
\textbf{2019} & 67  & 784 &  18  & 331   & 62  & 281 & 1543 \\
\textbf{2020} & 75  & 710 &  65  &  171   & 33   & 313 & 1367\\ \hline
\textbf{$\Sigma$} & 375  & 2840 &  218  & 1020   & 234  & 1162 & 5849\\

	\end{tabular}
	\label{tab:numberSessions}
\end{table}

\begin{figure}[htb]
\centering
\begin{minipage}{0.32 \linewidth}
 \includegraphics[width=\linewidth]{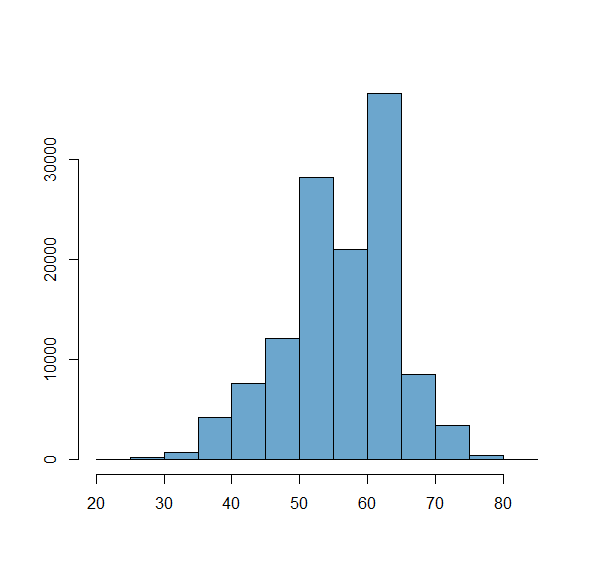} \vspace{-10mm} \\ \centerline{Bulgaria.}    
\end{minipage}
\begin{minipage}{0.32 \linewidth}
 \includegraphics[width=\linewidth]{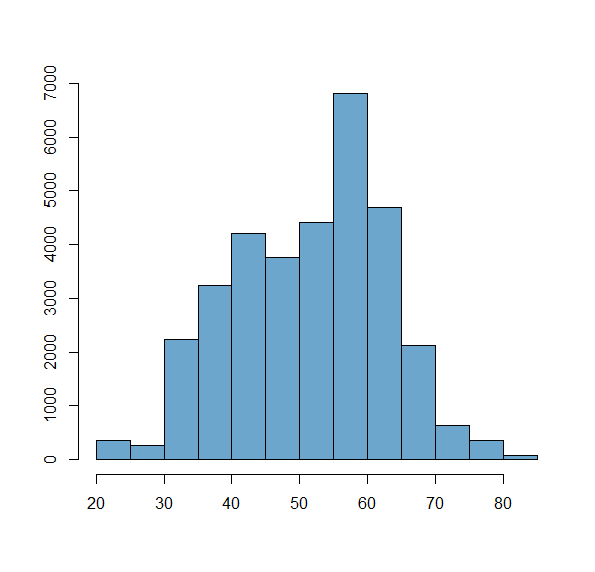} \vspace{-10mm} \\ \centerline{Czech Republic.}    
\end{minipage}
\begin{minipage}{0.32 \linewidth}
 \includegraphics[width=\linewidth]{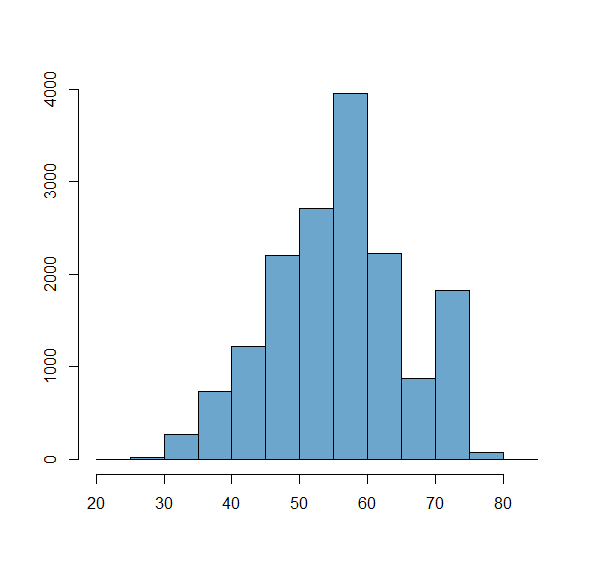} \vspace{-10mm} \\ \centerline{Spain.}    
\end{minipage}


\begin{minipage}{0.32 \linewidth}
 \includegraphics[width=\linewidth]{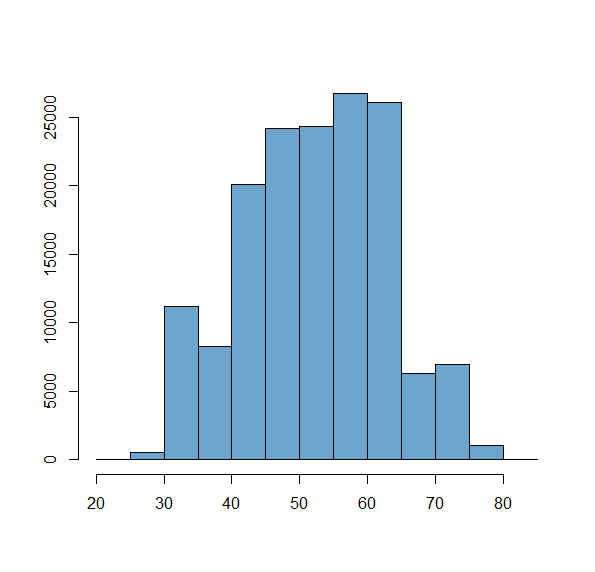} \vspace{-10mm} \\ \centerline{France.}    
\end{minipage}
\begin{minipage}{0.32 \linewidth}
 \includegraphics[width=\linewidth]{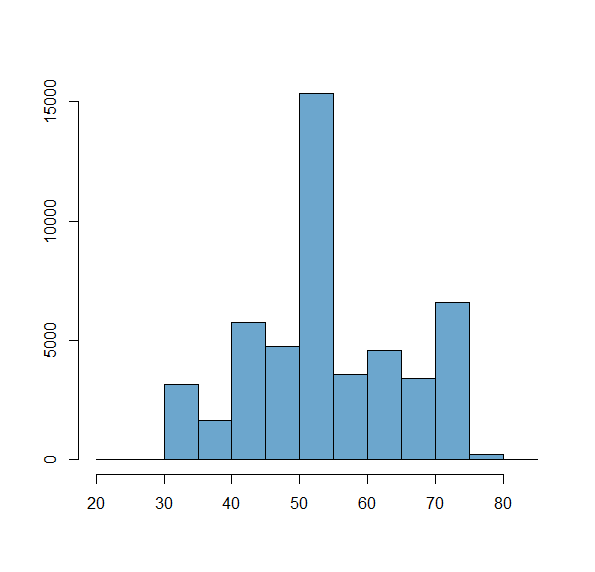} \vspace{-10mm} \\ \centerline{Slovenia.}    
\end{minipage}

\vspace{5mm}

\caption{The distribution of the number of speeches relative to the MP's age across the parliaments. The age information is not available for the UK.}
\label{fig:histoAge}
\end{figure}

\begin{figure}[!!htb]
  \centering
    \includegraphics[width=\linewidth,height=10cm]{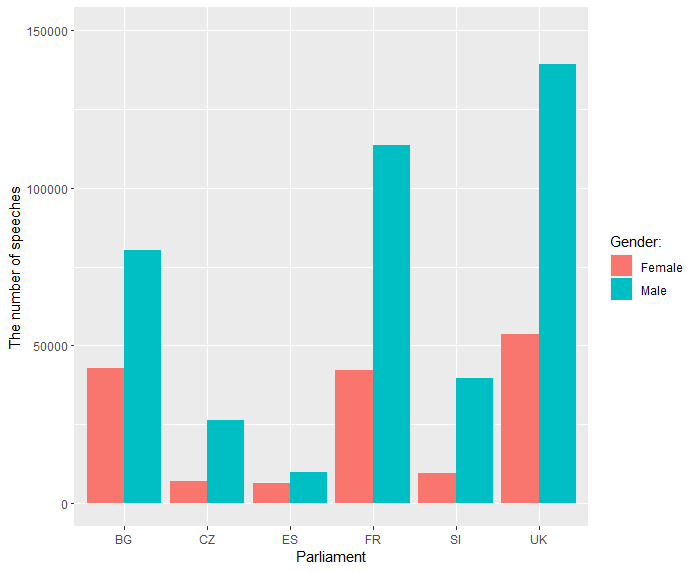}
        \caption{The distribution of the number of speeches relative to the MP's gender across the parliaments. }
        \label{fig:histoGender}
\end{figure}

We show the distribution of the number of speeches based on MP's age and gender across different parliaments in Figures \ref{fig:histoAge} and \ref{fig:histoGender}, respectively. To get representative age cohorts, we merge MPs into 5-year intervals for \ref{fig:histoAge}. The distributions based on speakers' age tend to peak between 50 and 60, except for France where there are no peaks. Concerning gender, there are considerably more speeches of male MPs in all parliaments, with the smallest difference in gender in Spain and the largest in Slovenia and Czechia.

\section{Methodology}
\label{sec:methodology}
We selected several modern NLP approaches to analyze the speeches in chosen languages. 
We use supervised text classification for sentiment and emotion analysis, as well as to predict the age, gender, and political wing of a speaker. Since 2019, a standard approach to text classification has been fine-tuning one of the large pre-trained language models such as BERT \cite{devlin-etal-2019-bert} to the specific task. We followed this approach and used multilingual BERT (pre-trained in 104 languages) to predict desired variables in a uniform way across the tackled languages. The trained models were used to predict speakers' meta-information (gender, age, political wing), as described in Section \ref{sec:metadataPrediction}, as well as their sentiments and emotions based on the language and contextual information in speeches, as outlined in Sections \ref{sec:sentimentPrediction} and \ref{methodology:emotion}, respectively. The details are presented below.

\subsection{Meta Data Prediction}
\label{sec:metadataPrediction}

Datasets from the ParlaMint project contain information about the speaker's age, gender, and political party. Thus, we fine-tune the multilingual BERT model to predict each of the three metadata variables from individual speeches of parliament members. The prediction accuracy of the models reveals the amount of information about the metadata stored in the parliament speeches. The variables that we predict are:

\begin{enumerate}

    \item Age of the speaker. The original corpora contained the birth year of the speaker, from which we computed the speaker's age. We dichotomized the age into two groups, separately for each country, using the Bayesaian AB testing to maximize the difference in sentiment between the younger and older group of MPs. The details of this process are contained in Section  \ref{results:sentiment_age}.

    \item Gender of the speaker. The original datasets contain a meta-data variable with the value 'F' for female and 'M' for male speakers, which we converted into 0 for females and 1 for males. We randomly selected \num{2500} speeches by male and \num{2500} speeches by female MPs. 
    
    \item Political wing of the speaker. We assumed that both left and right political parties can be split into moderate (center-leaning) and extreme (far-left and far-right). We separately predicted the differences between the moderate parties (i.e. center-left and center-right) and extreme parties (extreme-left and extreme-right). For both center-leaning and extreme party comparisons, we selected \num{2500} speeches from the left and \num{2500} speeches from the right-wing of the political spectrum. 
        
\end{enumerate}

Each of the created datasets was split into training and testing data parts (80 \% for training and 20 \% for testing). The training set was used to fine-tune the multilingual BERT (mBERT) model\footnote{\url{https://huggingface.co/bert-base-multilingual-cased}}. BERT \cite{devlin-etal-2019-bert} is a text representation model based on the transformer neural network architecture \cite{Vaswani2017}, pre-trained on the masked language modelling task using a large corpus of data. The mBERT is pre-trained on 104 languages. In our work, we used the pre-trained models available in the HuggingFace platform and fine-tuned them separately for each task.

\subsection{Sentiment Prediction}
\label{sec:sentimentPrediction}

For the automatic classification of sentiment in text data, various approaches have been developed, the most successful being machine learning classifiers trained on human-annotated corpora. The main challenge of these approaches is that they tend to be domain-specific and work best when trained with labelled data from the target domain but are less effective in other domains. However, as producing labelled datasets is expensive, researchers often apply the trained models across domains and languages. The cross-lingual transfer is possible either by the machine translation from a language without a suitable dataset to a language where such a dataset exists or by using a pre-trained multilingual language model such as mBERT. 

As there are no specific parliamentary language sentiment datasets, our cross-lingual and cross-domain approach relies on a collection of sentiment datasets from different languages in the domain of news and media. The reason to choose the news sentiment datasets is that the language and the context used are relatively similar to the parliament discourse. We use two-class sentiment prediction with the negative sentiment labelled 0 and the positive labelled 1. As our previous work has shown that sentiment classifiers can be successfully transferred  between languages (especially similar ones) with mBERT \cite{RobnikSikonja2021}, we combined datasets from three languages as follows.

\begin{enumerate}
    \item \textit{Slovenian SentiNews dataset} \cite{buvcar2018annotated} is a manually labelled sentiment dataset containing \num{10427} documents there were annotated on a document, paragraph and sentence level. For our task, we used the document level annotation selecting the negative and positive labelled news. The selected instances consist of \num{3337} negative and \num{1665} positive Slovene news. 
    \item \textit{English news headlines datasets}, consisting of two sources:
    \begin{itemize}
        \item The financial news headlines dataset \cite{malo2014good} was labelled with the sentiment from the perspective of a retail investor and constructed based on the human-annotated finance phrase bank. The data contained \num{604} negative and \num{1363} positive headlines.
        \item The SEN dataset \cite{baraniak2021dataset} is a recent human-labelled dataset for entity-level sentiment analysis of political news headlines. The dataset consists of \num{3819} human-labelled political news headlines from several major online media outlets in English and Polish. Each record contains a news headline, a named entity mentioned in the headline, and a human-annotated label (positive, neutral, or negative). The original SEN dataset package consists of two parts: SEN-en (English headlines that split into SEN-en-R and SEN-en-AMT), and SEN-pl (Polish headlines). The English dataset names are coming from the way the annotation process was done. For SEN-en-R, each headline-entity pair was annotated via the open-source annotation tool doccano\footnote{\url{https://github.com/doccano/doccano}} by at least 3 volunteer researchers while for the SEN-en-AMT the Amazon Mechanical Turk service was used. For our task, we selected only the labelled instances from the two English datasets that were annotated as negative and positive, ending with \num{982} negative and \num{456} positive instances.
    \end{itemize}
    
    \item \textit{Russian news dataset} obtained from the Kaggle\footnote{\url{https://www.kaggle.com/competitions/sentiment-analysis-in-russian/data}} that contains \num{8000} sentiments annotated news in the Russian language. From these, we selected \num{1434} negative and \num{2795} positively labelled instances.
\end{enumerate}
By combining all the above datasets, we obtained our final training dataset with \num{12636} labelled instances, of which there are \num{6357} negative and \num{6279} positive.

\subsection{Emotion Detection}
\label{methodology:emotion}

Besides informative content, texts also communicate attitudinal information, including emotional states \citep{alm2005emotions}. As there exist many emotional states and feelings, emotion detection is much more challenging compared to sentiment analysis, where typically three attitudes (positive, negative or neutral) are predicted from the given user input. 
In contrast to that, the emotion detection task deals with both primary emotions (e.g., happiness, sadness, anger, disgust, fear and surprise) as well as more complex emotion models involving different dimensions of emotions and psychology theories \citep{sailunaz2018emotion}. Apart from this, emotion analysis from texts suffers from relatively small and homogeneous annotated corpora \citep{cortal2022natural}. While there exist several English language datasets,  emotion analysis for less-resourced European languages is much more problematic \citep{oberlander2018analysis}. Most languages have very few if any, well-annotated datasets that can be used to train text classification models. To overcome this issue, we use multi-lingual and cross-lingual approaches and restrict
the covered emotions to positive and negative to get better statistical coverage and also reduce the error due to machine translations and/or cross-lingual transfer.

Similarly to the sentiment analysis, we use the mBERT model to detect emotions in a parliamentary speech. Our preliminary investigation showed that precise detection of many emotions is not possible in the multilingual setting, so we only categorized emotions into positive and negative. Again, the emotion detection datasets for the parliamentary domain and our languages (except English) do not exist, so we use several English language datasets from various domains to enable better generalization across domains. We fine-tune the mBERT model with the following four emotion-labelled datasets.
\begin{enumerate}
    \item The Kaggle Twitter dataset\footnote{\url{https://www.kaggle.com/datasets/pashupatigupta/emotion-detection-from-text}}  contains 13 different emotions and \num{40000} records. We selected \num{5209} happiness, \num{3842} love, \num{1323} hate, and \num{110} anger tweets. The instances were grouped into negative emotions (hate and anger within total \num{1433} instances) and positive emotions (happiness and love within total \num{9051} instances).
    \item  The HuggingFace\footnote{\url{https://huggingface.co/datasets/emotion}} Twitter dataset  \cite{saravia2018carer} contains \num{16000} annotated tweets. From these, we selected \num{1937} fear (labelled as negative) and \num{1304} love instances (labelled as positive).
\item GoEmotions\footnote{\url{https://ai.googleblog.com/2021/10/goemotions-dataset-for-fine-grained.html}} dataset \cite{demszky2020goemotions} is a human-annotated dataset of 58k Reddit comments extracted from popular English-language subreddits and labeled with 27 emotion categories. Some comments have multiple emotion labels but we selected only instances with a single labeled emotion. Extracted negative emotions are anger (\num{1025} instances) and disgust (\num{498} instances), while positive emotions are love (\num{1427} instances) and optimism (\num{861} instances). In total, we extracted \num{1523} negative and \num{2288} positive instances from this dataset.
    \item XE\footnote{\url{https://github.com/Helsinki-NLP/XED}} emotion dataset \cite{ohman2020xed} contains \num{25000} human-annotated Finnish and \num{30000} English sentences. From the English dataset, we selected anger and disgust sentences (in total \num{3803} instances) as negative emotions, and \num{1721} joy sentences as positive emotions.   
\end{enumerate}

Our final emotion detection dataset contains \num{23282} instances from which \num{8918} are labelled as containing negative and \num{14364} as expressing positive emotions. 

\section{Results}
\label{sec:results}
In this section, we report and interpret the obtained results. We present results related to the prediction of metadata (age, gender, and political wing) in Section \ref{results:metadata}, and sentiment and emotions analysis in Section \ref{results:sentiment}.

\subsection{Metadata Prediction: Age, Gender, and Political Wing}
\label{results:metadata}
As we described in Section \ref{sec:metadataPrediction}, we find-tuned the multilingual BERT language model to predict the speaker's metadata such as age, gender, and political position. The mBERT model was fine-tuned for each of the metadata variables and six countries separately, and we present the predictive performance measured on the testing datasets. Being able to predict any of these three variables indicates considerable differences in the language used by specific groups of parliamentary speakers. The differences and similarities between different countries are discussed below.

\subsubsection{Predicting the age of speakers}
\label{sec:resultsAge}
For each of the analysed parliaments, we split the speeches into three groups according to the first and the third age quartile. For each country, we created two prediction models, one trying to distinguish between the speeches of MPs younger than the first quartile and aged between the first and third quartile (Table \ref{tab:age1}), and the second distinguishing between the speakers aged between the first and third quartile and speakers older than the third quartile (Table \ref{tab:age2}). In this way, we investigate language differences between three generations of MPs, checking if their language is age-specific. 

Tables \ref{tab:age1} and \ref{tab:age2} show that age is a relatively well-predicted characteristic of speakers in Spain, Bulgaria, Slovenia and Czech Republic, while in France, there are very few language differences between speakers of different ages. The higher the prediction performance, the easier it is to distinguish between speakers' age groups, and the language generation gap is more significant. 

The only notable difference between Tables \ref{tab:age1} and \ref{tab:age2} are the scores for the Czech Republic. Here the gap between the youngest and middle-aged MPs is considerably larger than between the middle-aged and older MPs. We hypothesize that this might be the result of transition between the communist-rule and parliamentary democracy, where the language of younger generations is less affected by the previous social system.


\begin{table}[H]
	\caption{Classification accuracy, precision, recall, and $F_1$-score for predicting speakers' age (younger than  the first quartile or between the first and third quartile) in different parliaments. For the UK parliament, the age of speakers is not available.
	} 
	\centering
	\setlength{\tabcolsep}{6pt}
	\begin{tabular}{lcccccc}
	 \textbf{} &    \textbf{BG} & \textbf{CZ} &  \textbf{ES} &    \textbf{FR} & \textbf{SI} &   \textbf{UK} \\
		\hline			
\textbf{Accuracy} & 0.71 & 0.70 & 0.72 & 0.53  & 0.65 & /  \\
\textbf{Precision}& 0.76 & 0.71  & 0.74 & 0.56  & 0.66 & /  \\
\textbf{Recall}   & 0.64 & 0.69 & 0.77 & 0.43  & 0.65 & /  \\
\textbf{$F_1$}    & 0.69 & 0.70 & 0.76 & 0.49 & 0.66 & /  \\
\hline
	\end{tabular}
	\label{tab:age1}
\end{table}

\begin{table}[H]
	\caption{Classification accuracy, precision, recall, and $F_1$-score for predicting speakers' age (between the first and third quartile or older than the third quartile). 
	} 
	\centering
	\setlength{\tabcolsep}{6pt}
	\begin{tabular}{lcccccc}
	 \textbf{} &    \textbf{BG} & \textbf{CZ} &  \textbf{ES} &    \textbf{FR} & \textbf{SI} &   \textbf{UK} \\
		\hline			
\textbf{Accuracy} & 0.66 & 0.63  & 0.74 &  0.52  & 0.61 & /  \\
\textbf{Precision}& 0.67 & 0.64  & 0.77 & 0.53 & 0.63 & /  \\
\textbf{Recall}  &  0.65 & 0.61  & 0.75 & 0.54 & 0.61 & /  \\
\textbf{$F_1$}   &  0.66 &  0.63 & 0.76 & 0.53 & 0.62 & /  \\
\hline
	\end{tabular}
	\label{tab:age2}
\end{table}

Below we try to explain the two extreme cases, Spain with the largest gap between age groups and France with the smallest.

 \citet{flaherty1987langue} presents a historical development of French political discourse, which is directed toward uniformity in discursive strategies and may explain their similarities. A similar conclusion was drawn by \citet{lehti2014style} who show that the language used in French politicians' blogs is relatively standard. 
 
 The Spanish case, with the largest differences between younger and older parliamentary speakers, may be explained by the fact that after the end of the two-party system, new parties with younger leaders wanted to contrast with more senior and more socially privileged individuals \cite{cameron201110}.

\subsubsection{Predicting the gender of speakers}
As discussed in Section \ref{related:gender}, the information about speakers' gender may be detected from the grammatical structures used in their speech for all the analyzed languages but English if speakers use phrases related to their personal beliefs and feelings. Another possibility to detect gender is if speakers of different gender indeed use different languages. We test differences in the speech between genders in two ways: first, we predict the gender in the original language, and second, we translate all speeches to English, and predict the gender of the translated speeches, thereby avoiding possible leakage from grammatical structures in the original languages.  

As Table \ref{tab:gender} shows, gender in the original texts is detectable to some degree in all analyzed countries. Slavic language (Slovenian, Czech, and Bulgarian) speakers express their gender the most explicitly, followed by Spanish, English and French speakers. The last two (English and French) are surprising for different reasons. In French, where gender may be expressed with the language, there is little evidence that speakers express it. Similarly to age, we hypothesize that the case of French could be explained by the tendency toward language uniformity in French political discourse. 
Contrary to that, in English, where gender expression is not part of the grammar, the speakers' gender can be detected nevertheless, indicating differences in expression between male and female MPs. 

\begin{table}[H]
	\caption{Predictive performance of speakers' gender prediction for six \emph{original} parliamentary datasets.   
	} 
	\centering
	\renewcommand{\arraystretch}{1.2}
	\setlength{\tabcolsep}{6pt}
	\begin{tabular}{lcccccc}
	 \textbf{} &  \textbf{BG} &  \textbf{CZ}  &   \textbf{ES} &   \textbf{FR} &   \textbf{SI}  & \textbf{UK} \\
		\hline			
\textbf{Accuracy} & 0.70  & 0.86 &  0.66  & 0.58 & 0.88 & 0.58  \\
\textbf{Precision}& 0.69 & 0.87 &  0.66  & 0.56 & 0.92 & 0.59 \\
\textbf{Recall} &  0.72 & 0.85  &  0.71  & 0.60 & 0.93 & 0.64 \\
\textbf{$F_1$}  &  0.71 & 0.86  &  0.68 & 0.58 & 0.93 & 0.61  \\
\hline
	\end{tabular}
	\label{tab:gender}
\end{table}

In Table \ref{tab:genderTranslated} we see that gender differences are detectable also if the speeches are first translated to English (as gender neutral concerning grammar), indicating differences in the expression between male and female MPs. For Bulgarian, Spanish, and French, we can observe small differences in prediction scores compared to the original languages  (in Table \ref{tab:gender}), while Czech and Slovene show substantially lower prediction performance for the translated speeches. Nevertheless, in Slovene the gender differences are still the most pronounced. 

Based on both original and translated speeches, we can conclude that gender differences exist in all analyzed countries.

\begin{table}[H]
	\caption{Predictive performance of speakers' gender prediction for five parliamentary datasets \emph{translated into English}.
 For UK the translation makes no sense.
	} 
	\centering
	\renewcommand{\arraystretch}{1.2}
	\setlength{\tabcolsep}{6pt}
	\begin{tabular}{lcccccc}
	 \textbf{} &  \textbf{BG} &  \textbf{CZ}  &   \textbf{ES} &   \textbf{FR} &   \textbf{SI}  & \textbf{UK} \\
		\hline			
\textbf{Accuracy} & 0.73 & 0.65 & 0.66   & 0.58  & 0.77  & -   \\
\textbf{Precision}& 0.74 & 0.67 & 0.68  & 0.56  & 0.77  &  -\\
\textbf{Recall} & 0.74 & 0.62 & 0.65   &  0.54 & 0.79  &  -\\
\textbf{$F_1$}  & 0.74 & 0.64 & 0.67  & 0.55  & 0.78  &   - \\
\hline
	\end{tabular}
	\label{tab:genderTranslated}
\end{table}

\subsubsection{Predicting the political orientation of speakers}

This section investigates the speech differences between parliament members with different political orientations. Our approach is again based on prediction models that predict the metadata (party membership) available for speakers. A successful prediction would testify that speakers of different political orientations use different languages, while low success in prediction would indicate that the compared parties use similar discourse.
We investigate two scenarios of different difficulties:
\begin{enumerate}
    \item Predicting the left/right positioning of speakers from firmly or extreme left and right political parties. This problem shall not be very difficult, as we expect significant differences in the political stance between these parties, which we assume will be expressed in different content and possibly other linguistic features. The results are presented in Table \ref{tab:position-EXT}.
    \item Predicting the left/right positioning of speakers from the center-left and center-right political parties. This shall be a more complex problem as we try to distinguish between speakers from relatively similar parties. The results are presented in Table \ref{tab:position-CEN}.
\end{enumerate}


\begin{table}[H]
	\caption{Classification accuracy, precision, recall and $F_1$ score when predicting the political wing of speakers from the extreme left and extreme right parties in different parliaments. 
		} 
	\centering
	\renewcommand{\arraystretch}{1.2}
	\setlength{\tabcolsep}{3pt}
	\begin{tabular}{lcccccc}
	 \textbf{} & \textbf{BG} & \textbf{CZ} &  \textbf{ES} & \textbf{FR}  & \textbf{SI} &    \textbf{UK}  \\
		\hline			
\textbf{Acc} &  0.80  & 0.88  & 0.80  & 0.74  & 0.84 & 0.78  \\
\textbf{Pre} &  0.78  & 0.87  & 0.82  & 0.75  & 0.88  & 0.81  \\
\textbf{Rec} &  0.77  & 0.86  & 0.77  & 0.70 & 0.80  & 0.75  \\
\textbf{F1}  &  0.78  & 0.86  & 0.80  & 0.73  & 0.84  & 0.78  \\ \hline
			\end{tabular}
	\label{tab:position-EXT}
\end{table}

\begin{table}[H]
	\caption{Classification accuracy, precision, recall and $F_1$ score when predicting the political wing of speakers from center left and center right parties in different parliaments. 
		} 
	\centering
	\renewcommand{\arraystretch}{1.2}
	\setlength{\tabcolsep}{3pt}
	\begin{tabular}{lcccccc}
	 \textbf{} & \textbf{BG} & \textbf{CZ} & \textbf{ES} &  \textbf{FR}  & \textbf{SI} &    \textbf{UK}  \\
		\hline			
\textbf{Acc} &  0.78  & 0.78 & 0.70 & 0.58 & 0.84 & 0.76  \\
\textbf{Pre} &  0.79  & 0.8 & 0.69  & 0.62 & 0.86 & 0.78   \\
\textbf{Rec} &  0.78  & 0.73 & 0.79 & 0.48 & 0.89 & 0.74  \\
\textbf{F1}  &  0.79  & 0.76 & 0.74 & 0.54 & 0.87 & 0.76  \\ \hline
			\end{tabular}
	\label{tab:position-CEN}
\end{table}

As expected, the differences in speech between extreme left- and right-wing parties are relatively well predictable for all countries, indicating big differences in the discourse of these parties. The classification accuracy between countries ranges from 88\% (Czech Republic) to 74\% (France).

Surprisingly, the differences are still large between center-left and center-right parties, ranging from 87\% (Slovenia) to 54\% (France). France is an exception with its low predictability (again, likely due to the tendency for uniform political discourse), which is much more prominent in other countries. For two countries, Slovenia and Bulgaria, the differences between central parties are larger than between extreme parties, which may indicate strong political competition between the central parties.

\subsection{Sentiment and Emotions Detection}
\label{results:sentiment}

This section presents the results obtained from the sentiments and emotion detection experiments. For these experiments, we fine-tuned multilingual BERT on the training datasets described in Section \ref{sec:sentimentPrediction} (sentiment) and Section \ref{methodology:emotion} (emotions). First, we try to establish the quality of the trained models. For that purpose, during the fine-tuning process, a small part of the training data (10 \% of all training instances) was used for the validation after each training epoch. This classification accuracy is shown in Table \ref{table:training} for both the sentiment and emotion detection tasks. The results show that sentiment and emotions can be relatively well-predicted, which is a positive indication of the reported results' reliability. Slightly better results in predicting positive and negative emotions are expected, as emotion datasets are all in English, all collected from social media, and therefore relatively homogeneous. The sentiment detection datasets are multilingual and collected in different domains; thus, the training accuracy reported in Table \ref{table:training} is expected to be lower. However, this does not mean that the obtained sentiment model would provide less good generalization on our out-of-domain parliamentary data.

\begin{table}[H]
\caption{Classification performance on the validation sets during fine-tuning the mBERT model on sentiment and emotion prediction tasks.}
\renewcommand{\arraystretch}{1}
\setlength{\tabcolsep}{5pt}
\label{table:training}
\centering
\begin{tabular}{ccc}

    Epoch     & \textbf{Sentiment} & \textbf{Emotions}   \\
    \hline
    1 & 0.85 & 0.90  \\
    2 & 0.86 & 0.91  \\
    3 & 0.87 & 0.92  \\
    4 & 0.87 & 0.92  \\
     \hline
\end{tabular}
\end{table}
 
 To further assess the quality of the produced sentiment prediction model, we selected 20 talks with the highest probability of the negative sentiment for each of the parliaments and manually validated weather predictions are correct. The results are presented in Table \ref{table:validationSen}. Based on the results, we consider the model's accuracy good enough (and comparable to other sentiment prediction models in the literature) to provide a reliable picture of the sentiment in our study. The fine-tuned models were used on our six parliamentary speech datasets. To obtain reliable and comparable statistics, we randomly selected \num{10000} speeches that have more than 30 characters of regular parliament members from 2020 
 for each of the six parliaments. For each speech, the trained mBERT model returned the sentiment score between 0 and 1 (0 indicating the negative and 1 indicating the positive sentiment).

\begin{table}[htb]
\caption{Manually determined percentage of instances with correctly predicted negative sentiment in parliamentary speeches for compared parliaments. 
}
\setlength{\tabcolsep}{5pt}
\label{table:validationSen}
\centering
\begin{tabular}{lc}
\textbf{Parliament}         & \textbf{Accuracy}  \\
    \hline
    \textbf{BG} & 85 \%\\
    \textbf{CZ} & 80 \%\\
    \textbf{ES} & 95 \%\\
    \textbf{FR} & 90 \%\\
    \textbf{SI} & 95 \%\\
    \textbf{UK} & 95 \%\\
     \hline
\end{tabular}
\end{table}

\begin{figure}[htb]
BG
\begin{minipage}{0.45 \linewidth}
 \includegraphics[width=\linewidth]{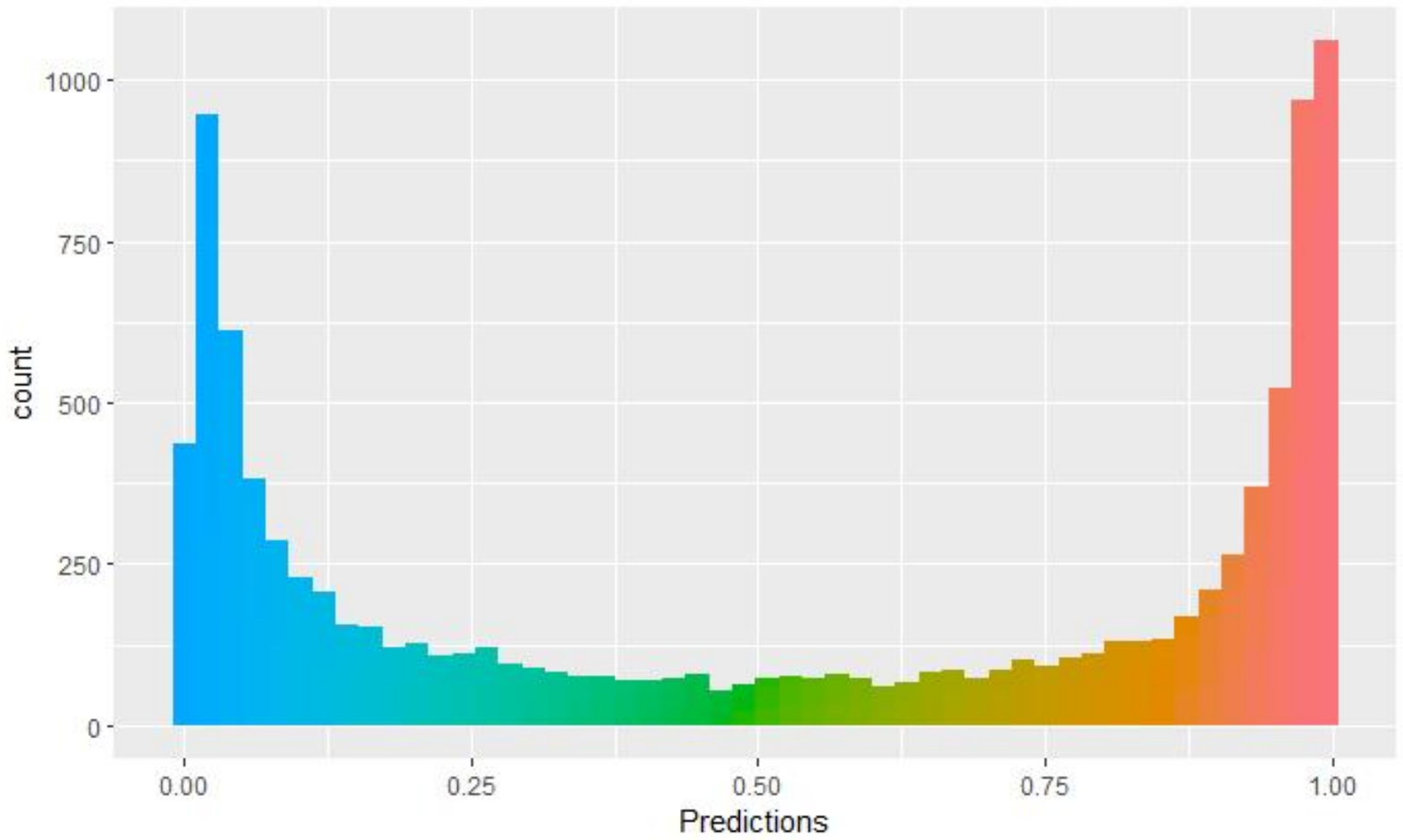}     
\end{minipage}
CZ
\begin{minipage}{0.45 \linewidth}
 \includegraphics[width=\linewidth]{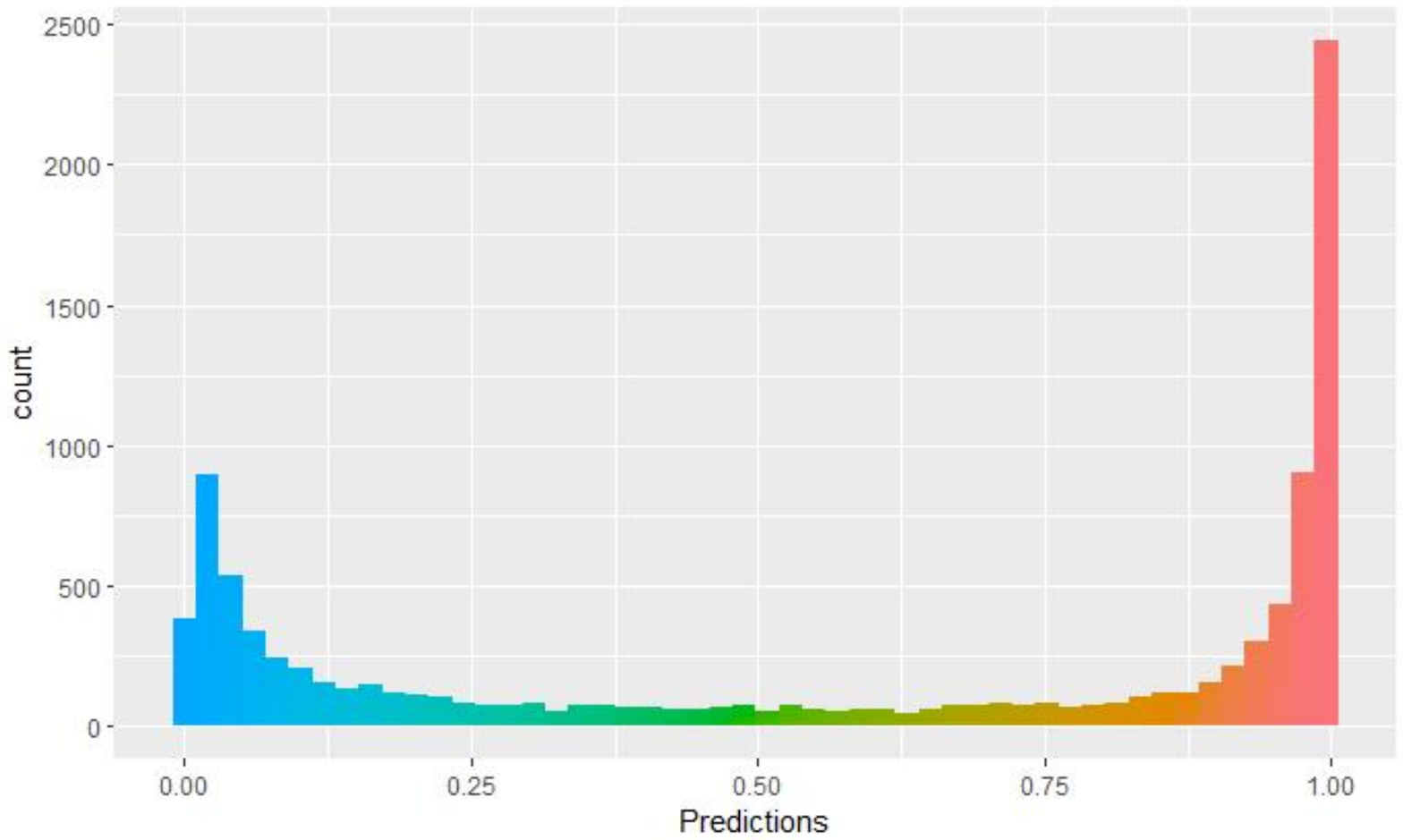}     
\end{minipage}

FR
\begin{minipage}{0.45 \linewidth}
 \includegraphics[width=\linewidth]{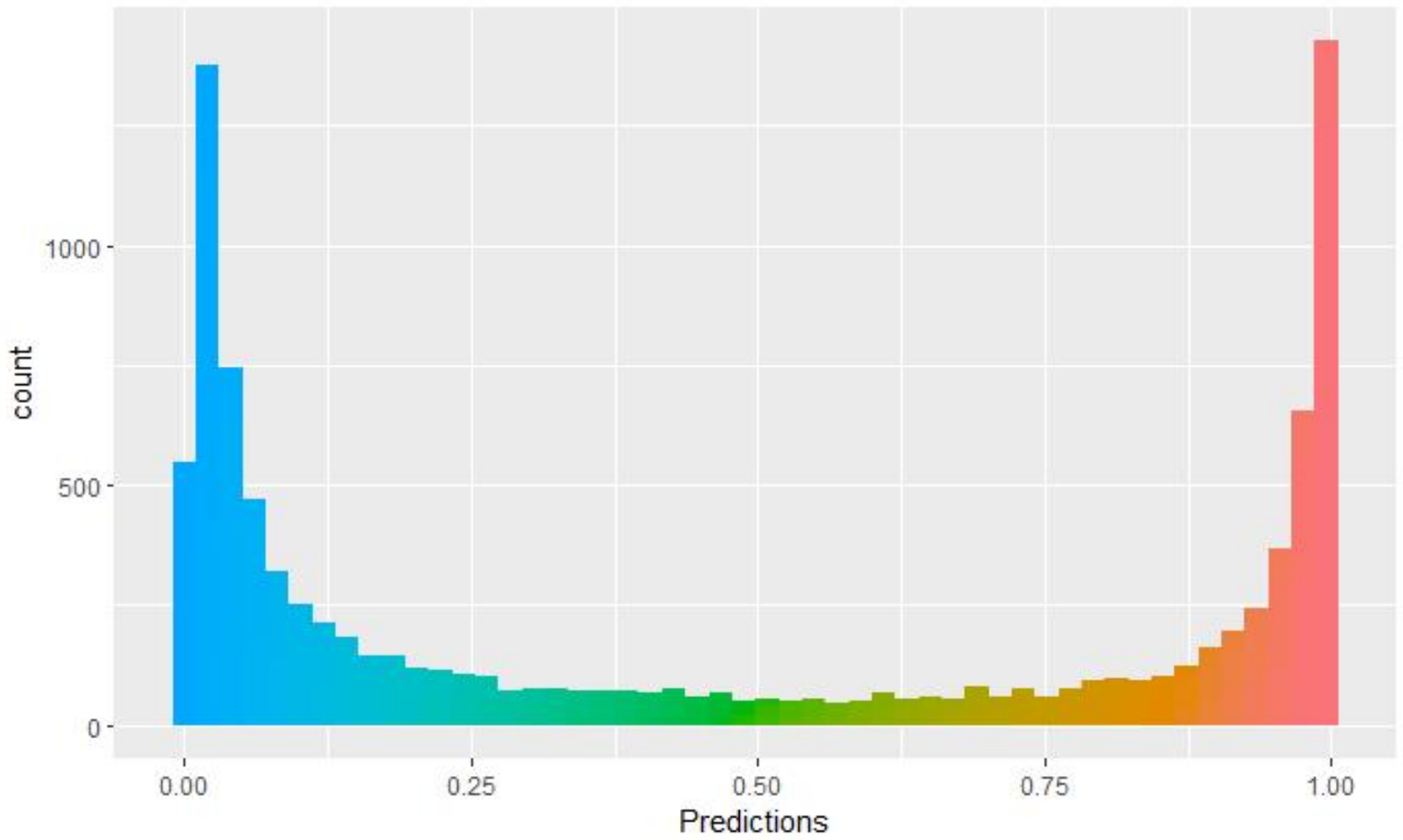}     
\end{minipage}
SI
\begin{minipage}{0.45 \linewidth}
 \includegraphics[width=\linewidth]{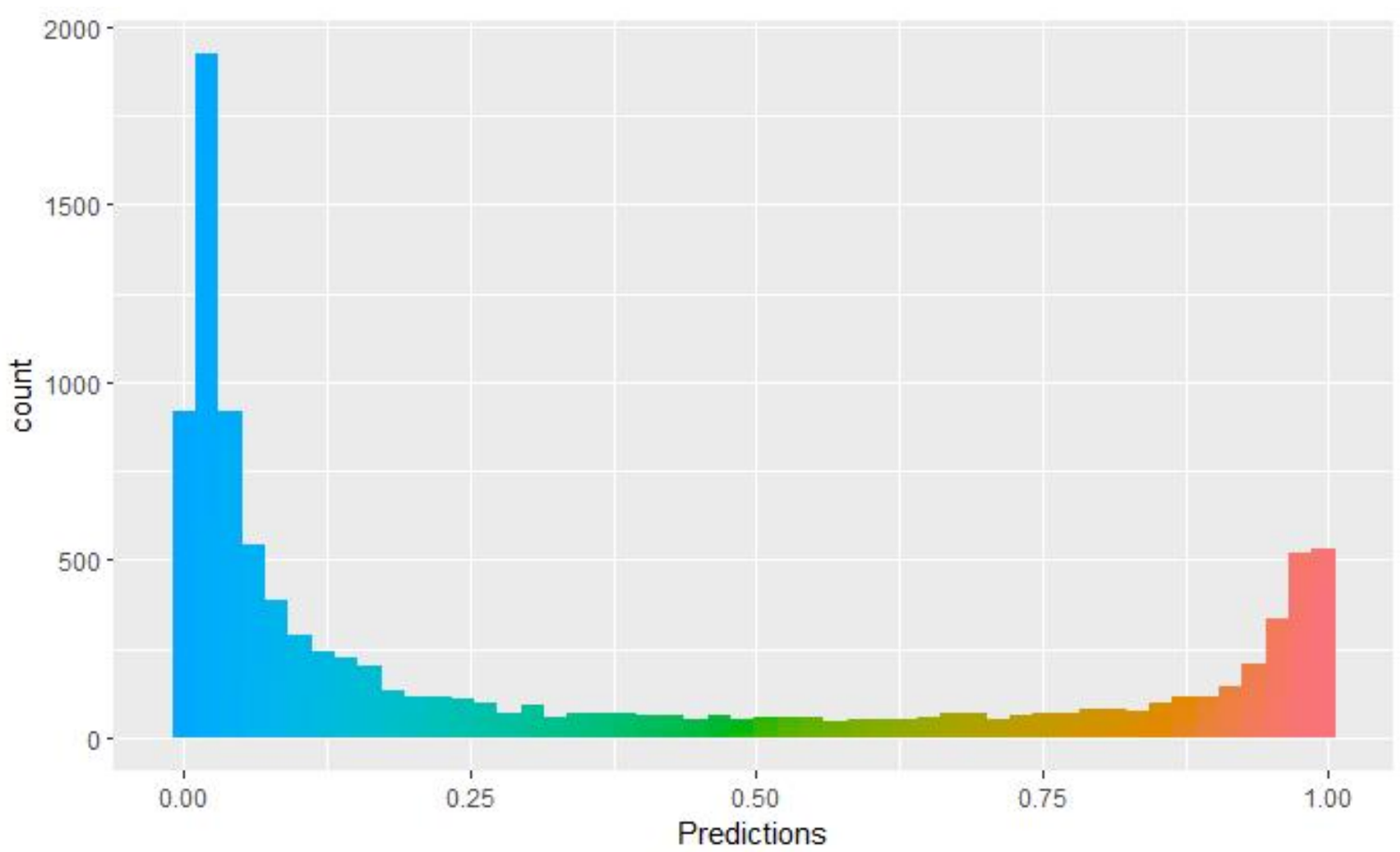}     
\end{minipage}

ES
\begin{minipage}{0.45 \linewidth}
 \includegraphics[width=\linewidth]{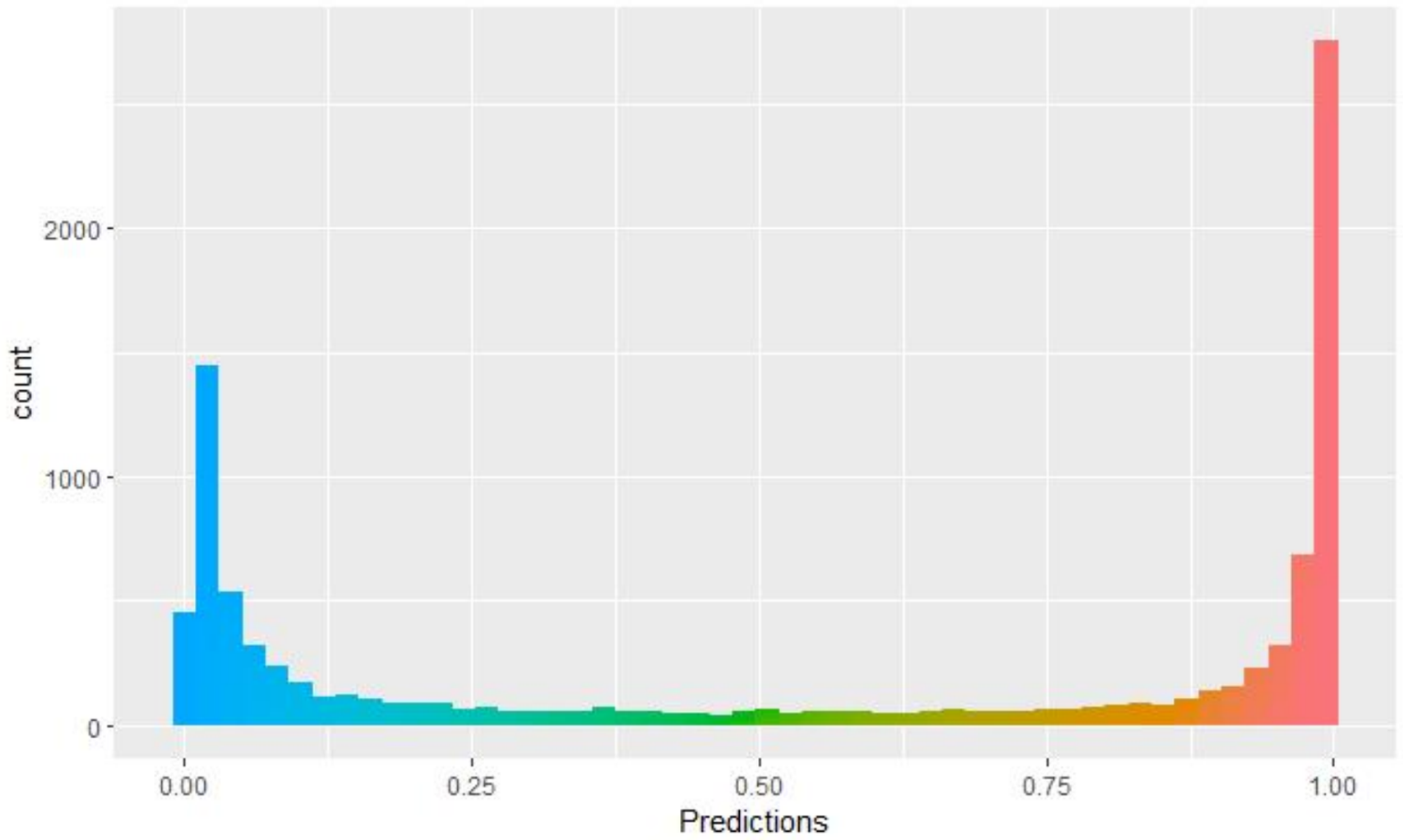}     
\end{minipage}
UK
\begin{minipage}{0.45 \linewidth}
 \includegraphics[width=\linewidth]{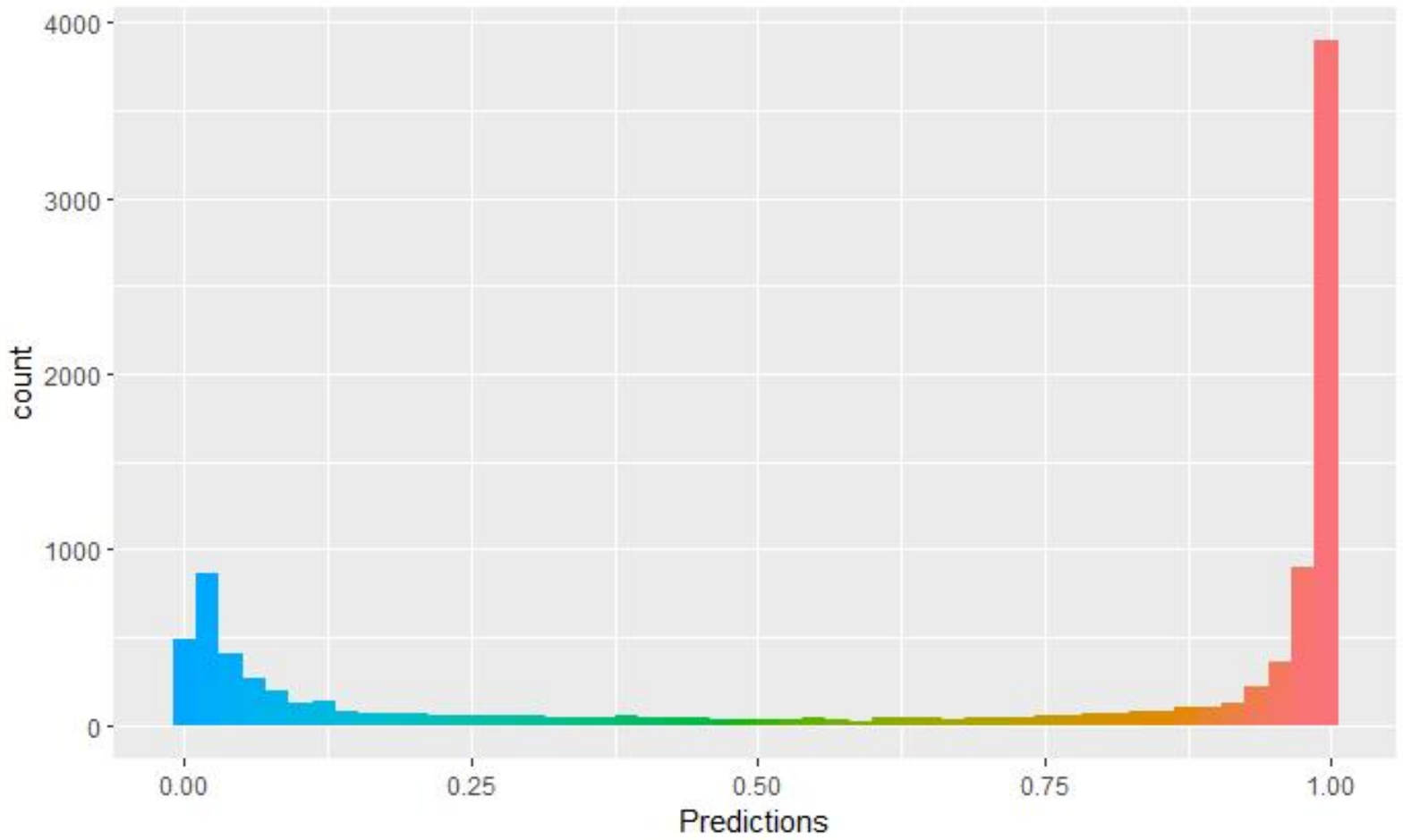}s     
\end{minipage}

\caption{The histograms show the distribution of sentiment predictions for six parliaments. The score of 0 indicates completely negative and the score of 1 completely positive sentiment.}
\label{fig:histogramsSen}
\end{figure}

Predictions are summarized in Figure \ref{fig:histogramsSen} and Table \ref{table:sen} from which we can compare the parliamentary sentiment across the countries.  Figure \ref{fig:histogramsSen} shows the histogram of sentiment distribution in each country. The Czech, Spanish and United Kingdom parliaments seem to express less negative sentiment than positive; in the Bulgarian and French parliaments, there seems to be a relatively balanced situation, while the Slovenian parliament shows the least positive sentiment. We attribute the results for Slovenia to the poisonous exchanges between the pro-government and opposition parties at the observed time when the previous opposition took over the government in the middle of the mandate. To get a numeric overview of the sentiment, we set the decision threshold for negative sentiment at 0.2 and for positive sentiment at 0.8 and counted the number of negative and positive speeches. The results are presented in Table \ref{table:sen}. As before, we conclude that the parliament with the highest percentage of negative sentiment is Slovenian, while the UK parliament speeches contain the highest positive sentiment rate.

\begin{table}[htb]
\caption{Percentage of negative and positive sentiment in parliamentary speeches for observed countries.}
\renewcommand{\arraystretch}{1}
\setlength{\tabcolsep}{5pt}
\label{table:sen}
\centering
\begin{tabular}{lcc}

         & \textbf{Negative} & \textbf{Positive}   \\
Parliament      & \textbf{sentiment} & \textbf{sentiment}   \\
    \hline
    \textbf{BG} & 35.67 & 39.68  \\
    \textbf{CZ} & 31.66 & 48.58  \\
    \textbf{FR} & 44.34 & 34.75  \\
    \textbf{SI} &\textbf{58.06} & 22.14  \\
    \textbf{ES} & 36.32 & 46.41  \\
    \textbf{UK} & 27.34 & \textbf{59.52}  \\
     \hline
\end{tabular}
\end{table}

Similarly to sentiment, we process emotions. To validate how good the emotions detection models are, we selected 20 speeches predicted to be the most negative 
for each parliament and manually checked if predictions were correct for them. The results are presented in Table \ref{table:validationEmo}. We can observe significantly lower accuracy in all countries compared to the sentiment (shown in Table \ref{table:validationSen}). While this is not surprising as the emotion prediction is considered harder compared to sentiment, this makes the results and interpretations presented below less reliable compared to the sentiment.

\begin{table}[hbt]
\caption{Manually determined percentage of instances with correctly predicted negative emotions in the parliamentary speeches of compared parliaments. 
}
\setlength{\tabcolsep}{5pt}
\label{table:validationEmo}
\centering
\begin{tabular}{lc}
\textbf{Parliament}         & \textbf{Accuracy}  \\
    \hline
    \textbf{BG} &  45 \%\\
    \textbf{CZ} &  40 \% \\
    \textbf{ES} &  65 \% \\
    \textbf{FR} &  55 \% \\
    \textbf{SI} &  55\% \\
    \textbf{UK} &  50 \% \\
     \hline
\end{tabular}
\end{table}

\begin{figure}[htb]
BG
\begin{minipage}{0.45 \linewidth}
 \includegraphics[width=\linewidth]{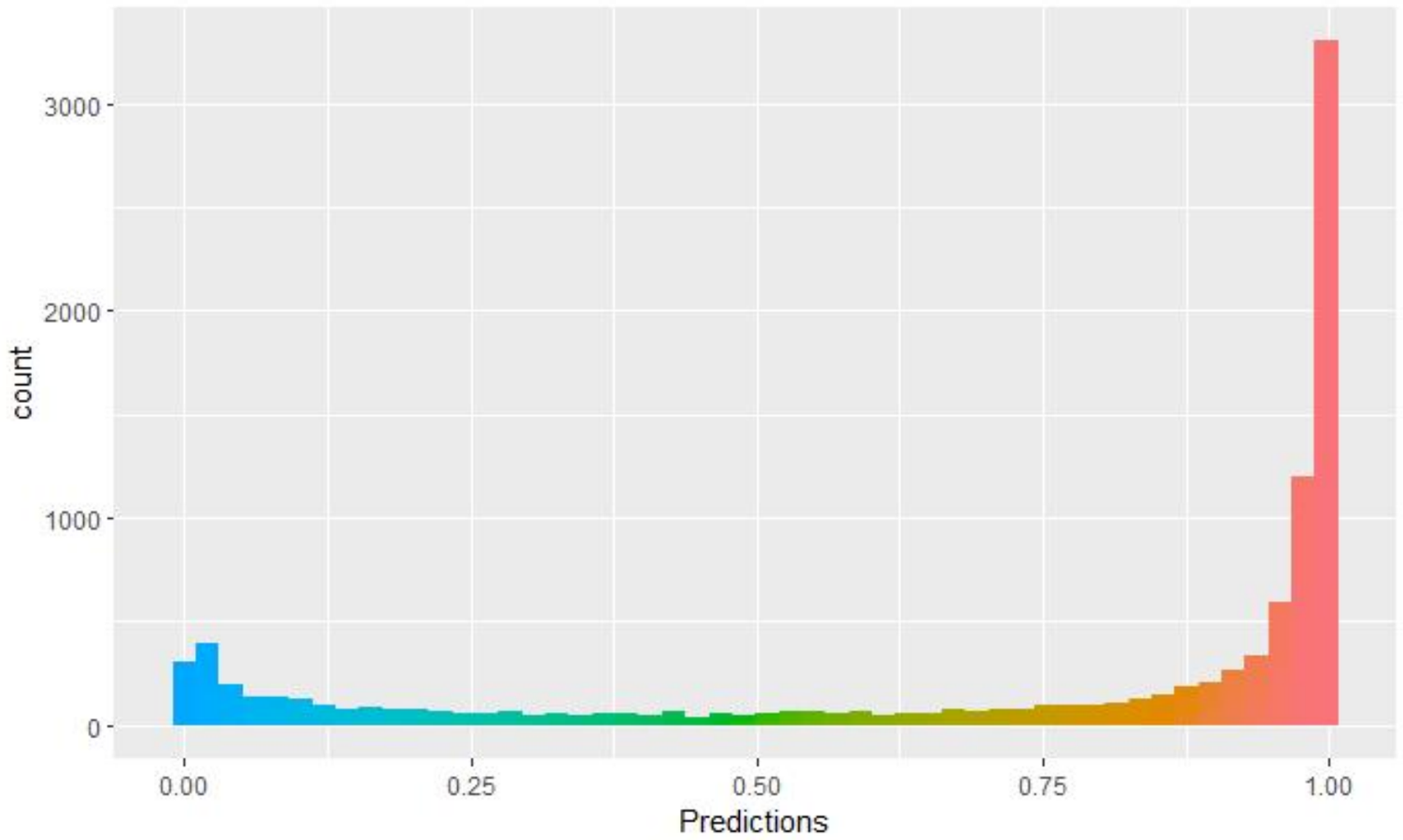}     
\end{minipage}
CZ
\begin{minipage}{0.45 \linewidth}
 \includegraphics[width=\linewidth]{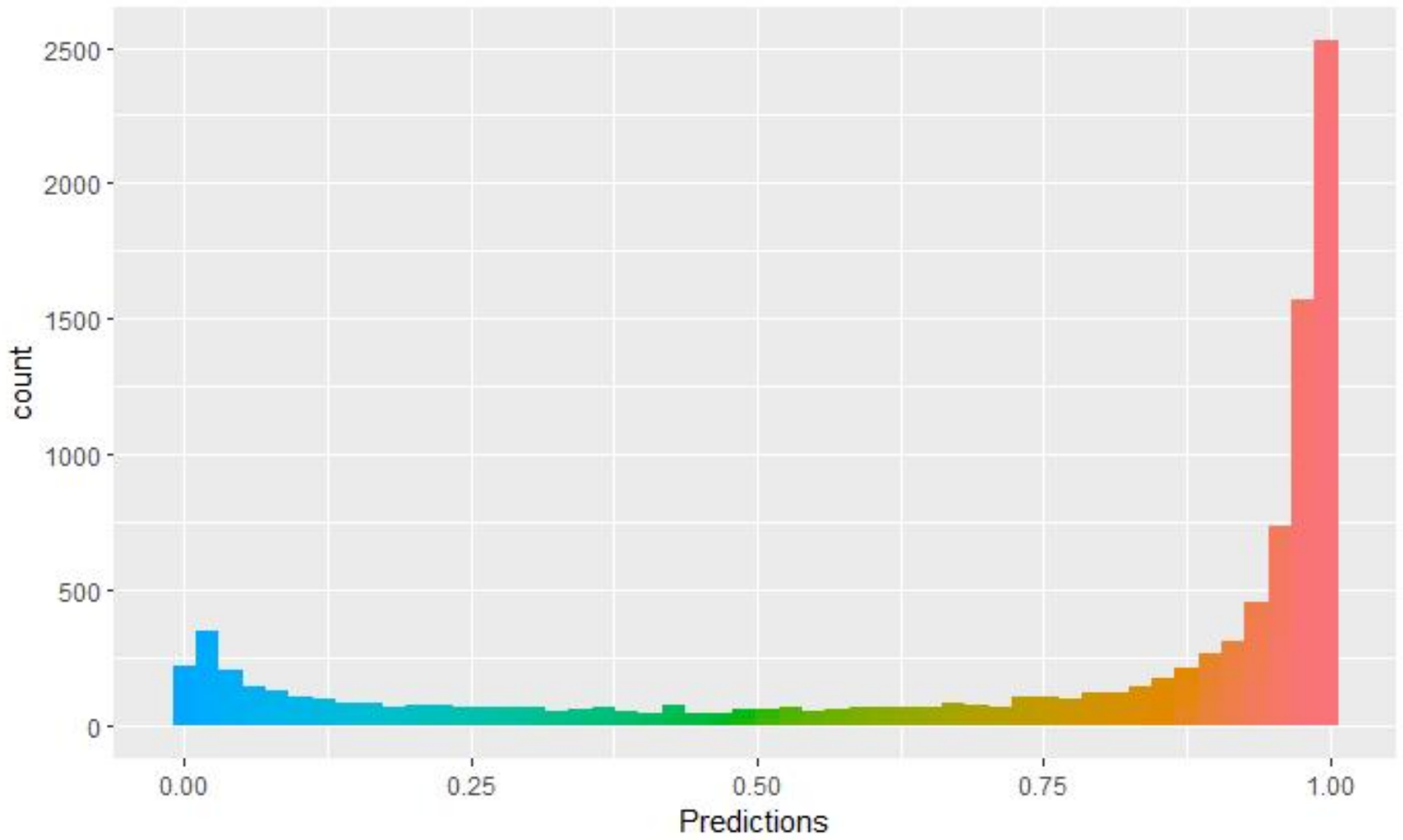}     
\end{minipage}

FR
\begin{minipage}{0.45 \linewidth}
 \includegraphics[width=\linewidth]{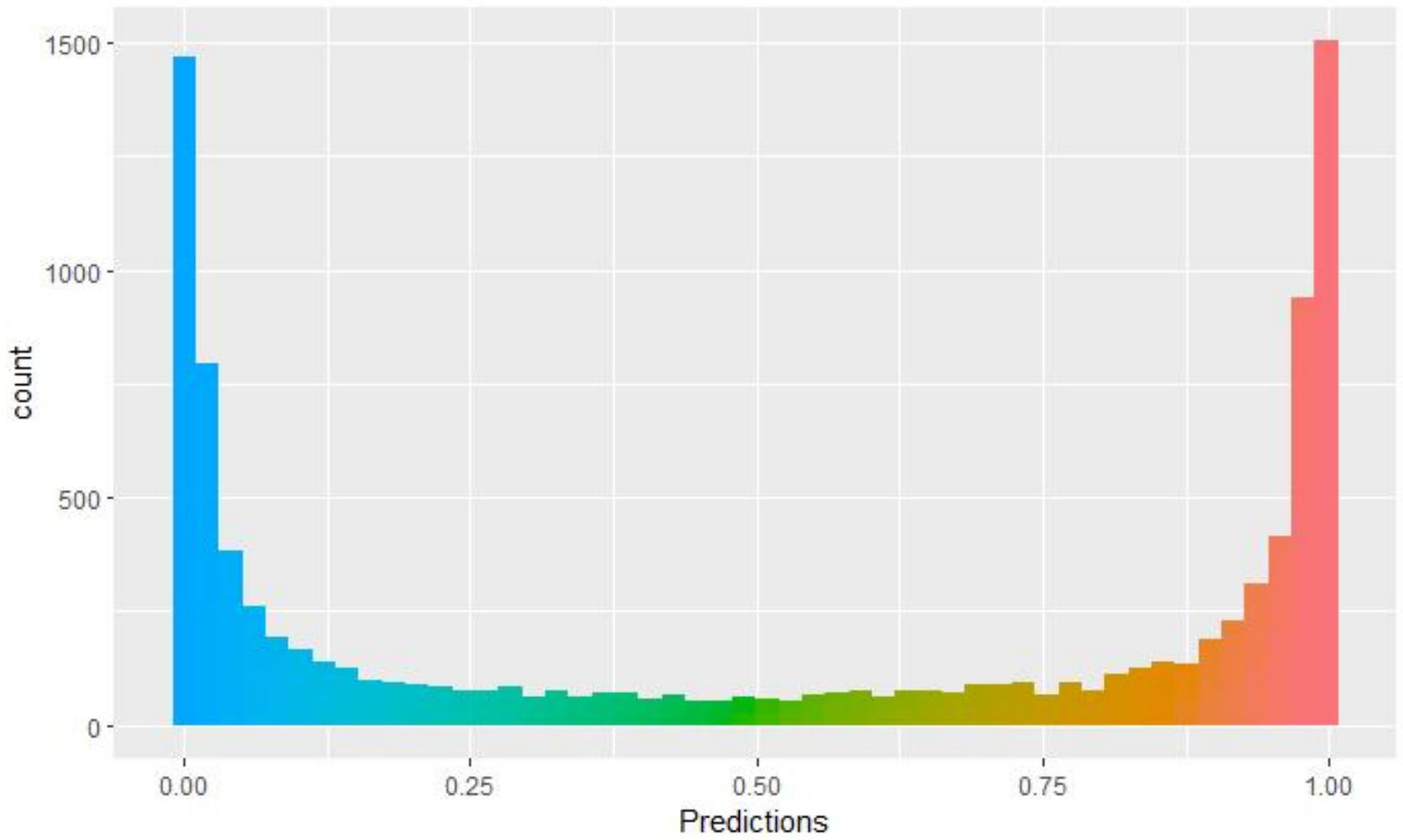}     
\end{minipage}
SI
\begin{minipage}{0.45 \linewidth}
 \includegraphics[width=\linewidth]{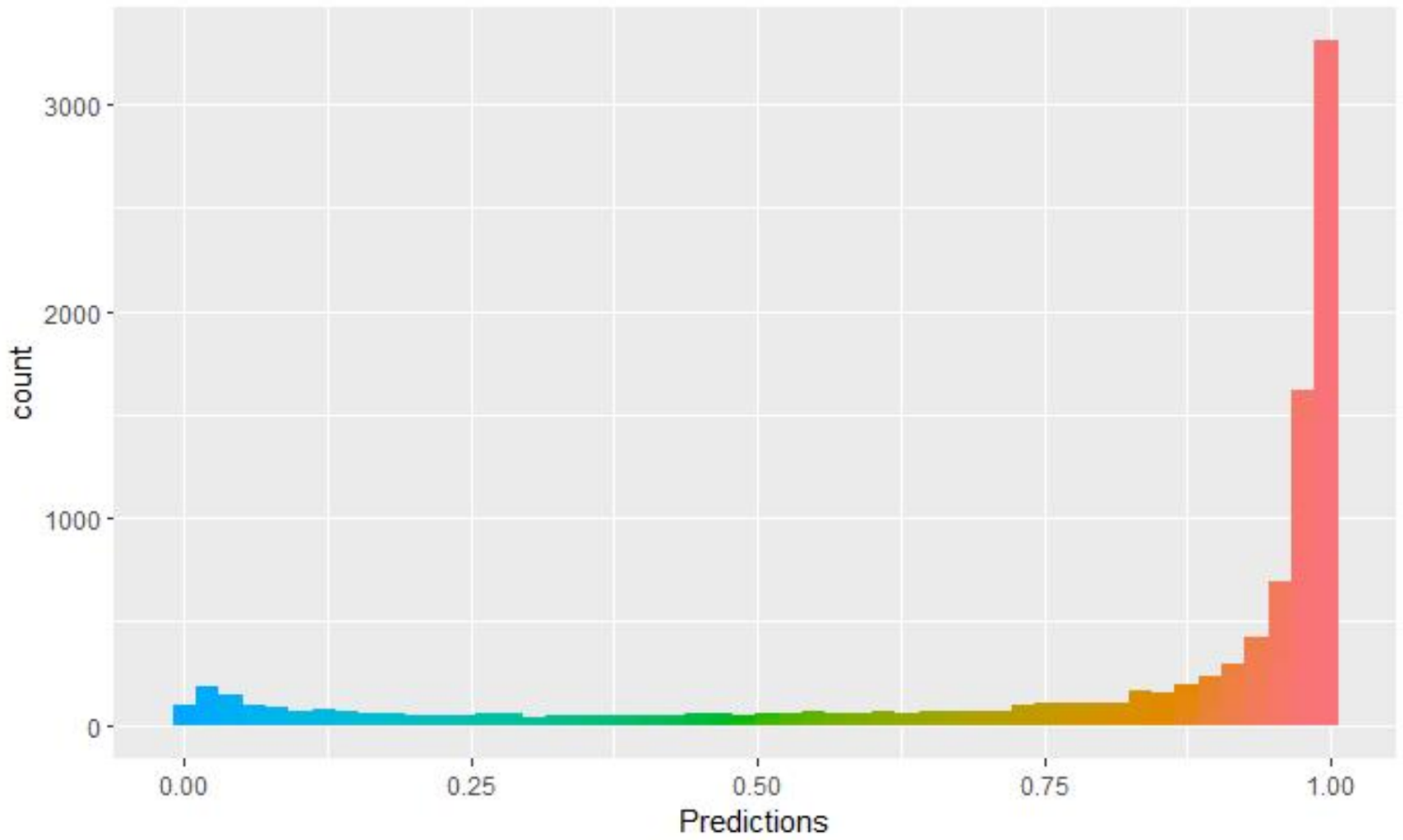}     
\end{minipage}

ES
\begin{minipage}{0.45 \linewidth}
 \includegraphics[width=\linewidth]{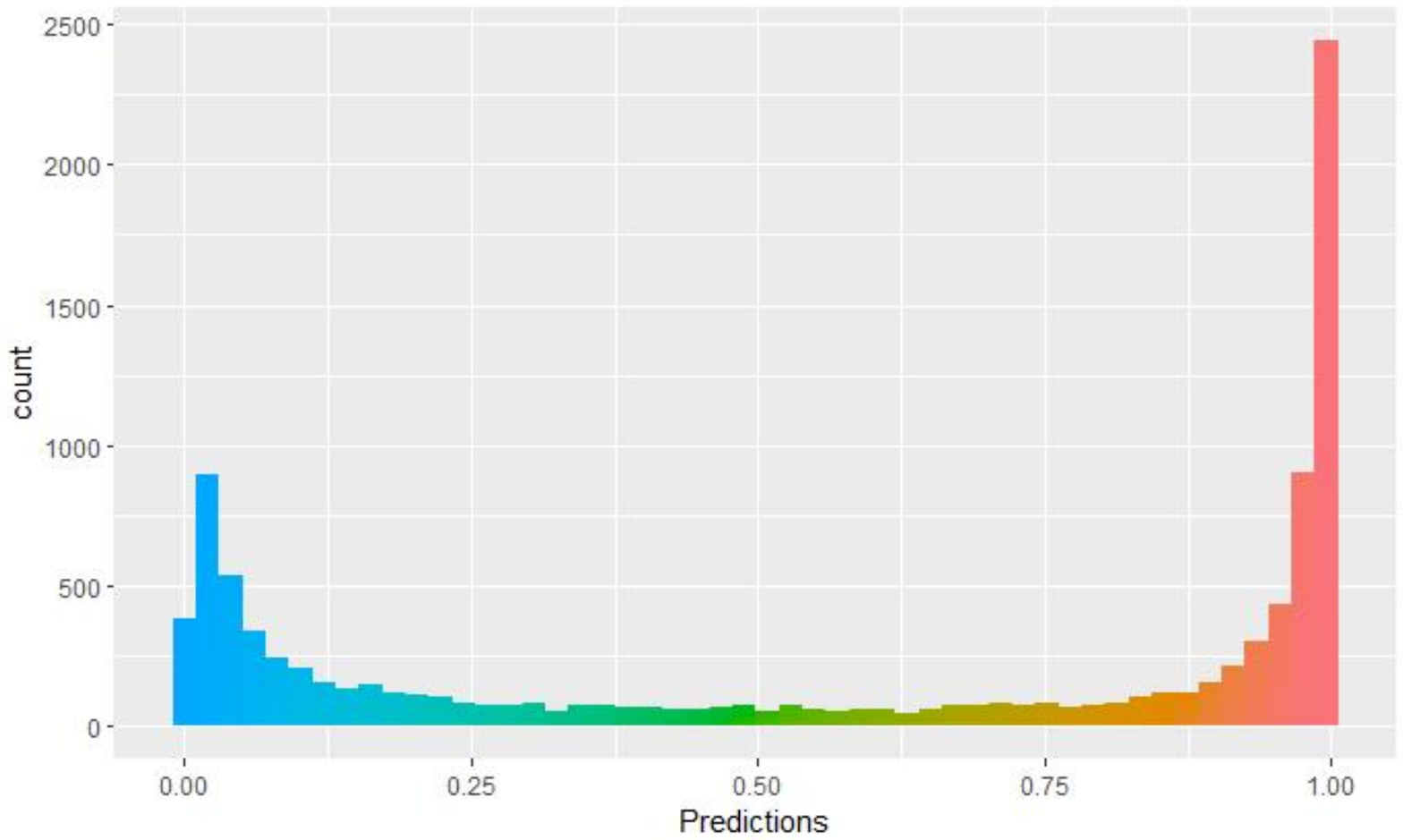}     
\end{minipage}
UK
\begin{minipage}{0.45 \linewidth}
 \includegraphics[width=\linewidth]{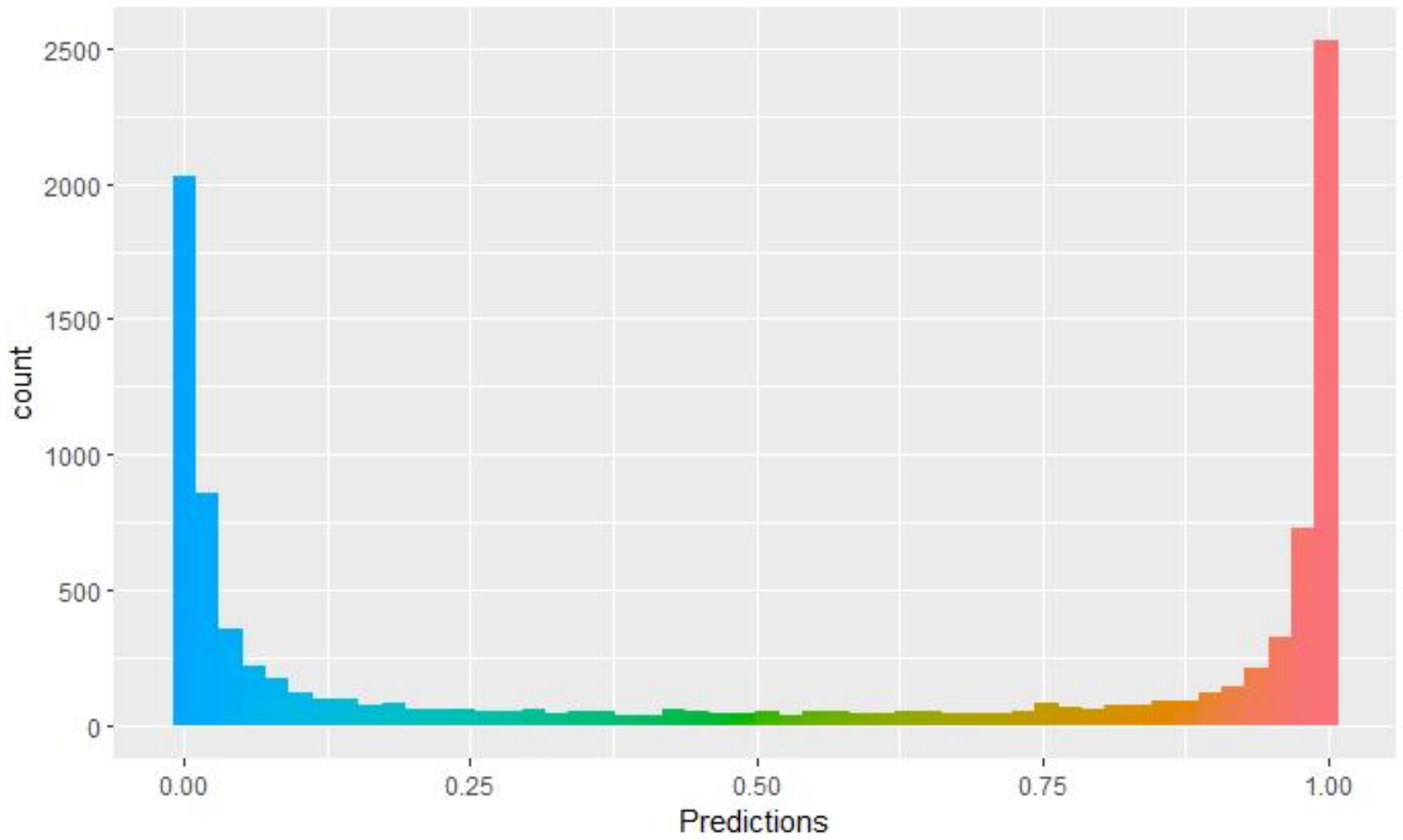}     
\end{minipage}

\caption{The distributions of emotion predictions for six parliaments. The score of 0 indicates completely negative emotions and the score of 1 completely positive emotions.}
\label{fig:histogramsEmo}
\end{figure}

We show the results for the emotion detection in Figure \ref{fig:histogramsEmo} (distribution of sentiment predictions) and Table \ref{table:emo} (the percentage of positive and negative emotions, taking 0.2 and 0.8 as the decision threshold values). As the results show, positive emotions are strongly dominant in all countries except France and UK, where positive and negative emotions are almost balanced.

\begin{table}[hbt]
\caption{Percentage of negative and positive emotions in parliamentary speeches for observed countries.}
\renewcommand{\arraystretch}{1}
\setlength{\tabcolsep}{5pt}
\label{table:emo}
\centering
\begin{tabular}{lcc}

         & \textbf{Negative} & \textbf{Positive}   \\
Parliament      & \textbf{emotions} & \textbf{emotions}   \\
    \hline
    \textbf{BG} & 16.5 & 64.86  \\
    \textbf{CZ} & 14.75 & 65.27  \\
    \textbf{FR} & 37.6 & 41.18  \\
    \textbf{SI} & 9.71 & 72.16  \\
    \textbf{ES} & 20.24 & 62.31  \\
    \textbf{UK} & 41.18 & 44.03  \\
     \hline
\end{tabular}
\end{table}

\subsection{The Impact of Age on Sentiment}
\label{results:sentiment_age}

In this section, we investigate the impact of age on the sentiment. We apply Bayesian statistics to find the age which best distinguishes between the positive and negative sentiments of the speakers.

We start with the hypothesis that older speakers more openly express the negative sentiments compared to younger speakers (the reverse hypothesis would be equally suitable for our approach), and form two hypotheses to use in the Bayesian hypothesis test:  
\begin{center}
    $H_0$: Younger MPs express the same amount of negative sentiment as the older ones. \\
    $H_1$: Younger MPs express more positive sentiments as the older ones.
\end{center}

To determine the age which best separates the younger from the older MPs in terms of positive sentiments, we estimated the posterior distribution for multiple age cutoff points as shown in Table \ref{table:BayesEmo}. The resulting numbers were constructed as follows:
\begin{enumerate}
    \item We separated the speakers to the younger and older based on the \textit{Age Cutoff}.
    \item For both, the younger and older population, we dichotomized the sentiment scores to 1 (scores higher than 0.5) or 0 (scores lower than 0.5), and assumed that they are drawn from the binomial distribution. Assuming the beta prior and the binomial likelihood, it is possible to estimate the closed-form Bernoulli posterior distributions.
    \item Using the \textit{bayesAB} R package\footnote{\href{https://CRAN.R-project.org/package=bayesAB}{https://CRAN.R-project.org/package=bayesAB}}, we estimated the posterior distributions for each of the two populations (younger and older) and calculated the probability that the sentiment scores of younger MPs are higher than the scores of older MPs. 
\end{enumerate}
The results show certain differences between the countries, In Bulgaria, Spain, and France, the MPs between 50 and 65 express the most positive sentiment, i.e. the thresholds for 55, 60, and 65 show high probabilities that younger MPs express positive sentiment while for the threshold of 50, this probability is lower. In the Czech Republic the positive sentiment is prominent in the age group of less than 55. In Slovenia, the MPs between 50 and 55 are the most negative, while other age groups are predominantly positive. 

\begin{table}[htb]
\caption{The probabilities that MPs younger than the Age Cutoff express more positive sentiments compared to older ones, as calculated with the Bayesian AB test. For UK parliament the age of MPs is not available.}
\setlength{\tabcolsep}{5pt}
\label{table:BayesEmo}
\centering
\begin{tabular}{ccccccc}
\textbf{Age Cutoff}         & \textbf{BG} & \textbf{CZ} & \textbf{ES} & \textbf{FR} & \textbf{SI} & \textbf{UK} \\
    \hline
    \textbf{50} & 61.2 & 83.8 & 27.3 & 16.5 & 88.1  & /\\
    \textbf{55} & 91.1 & 98.8 & 73.6 & 73.6 & 27.0 & /\\
    \textbf{60} & 95.5 & 54.9 & 100  & 95.6 & 94.1 & /\\
    \textbf{65} & 99.9 & 4.9 & 97.1 & 87.0 & 99.9  & /\\
    \textbf{70} & 73.7 & 0.8 & 7.8  & 73.3 & 94.1 & /\\
    \textbf{75} & 64.9 & 7.1 & 2.6  & 15.6 & 74.1 & /\\
     \hline
\end{tabular}
\end{table}





\subsection{Differences in political wing distributions}
\label{sec:discusion}
In this section, we further discuss some interesting questions related to the language of extreme left and right-wing politicians, i.e. we try to discover the language differences in political orientation in relation to age and gender. 
As explained in Section \ref{sec:metadataPrediction}, we separated political parties according to their political orientation (extreme left or extreme right), based on the publicly available information and parties' self-declarations.  

\begin{figure}[!!htb]
  \centering
\begin{minipage}{0.45 \linewidth}
    \includegraphics[width=\linewidth]{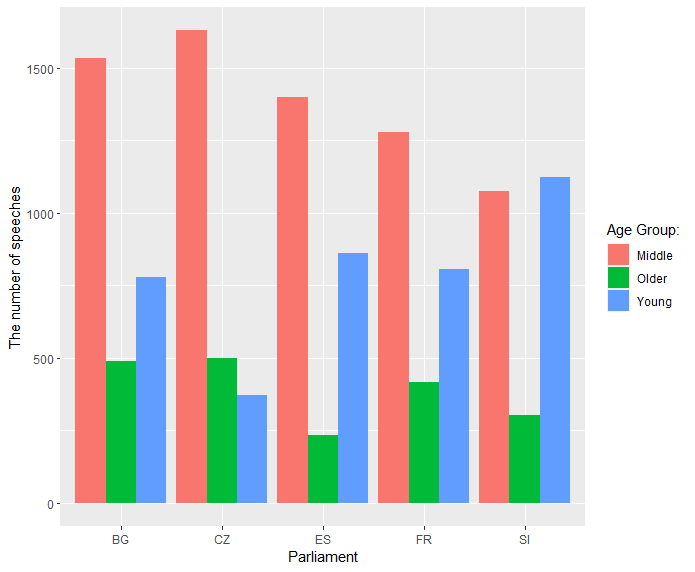} \\ \centerline{Extreme Left-wing speakers}
\end{minipage}
\begin{minipage}{0.45 \linewidth}
    \includegraphics[width=\linewidth]{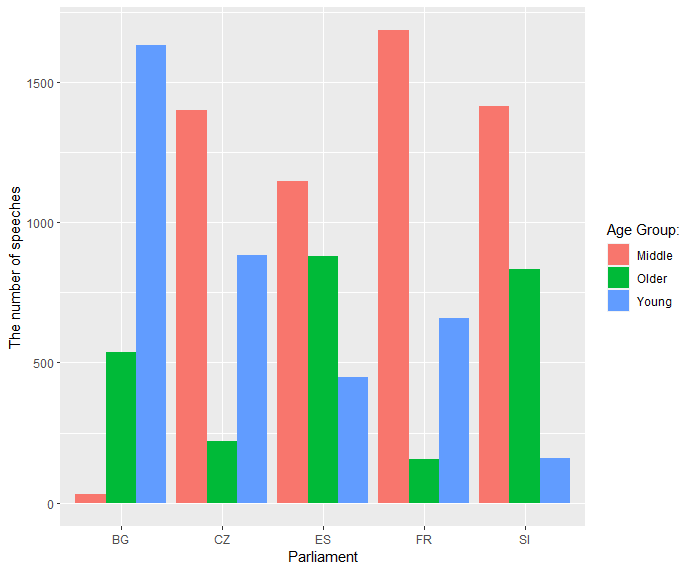} \\ \centerline{Extreme Right-wing speakers}
\end{minipage}
\caption{The distribution of the number of speeches relative to the MP's age group across the parliaments for extreme left- and right-wing MPs. }
        \label{fig:WingAge}
\end{figure}

We first plot the differences in political orientation based on the age of speakers. Similarly to Section \ref{sec:resultsAge}, we split the age span into three intervals, using the first and third quartile as the thresholds: younger MPs (less than the first quartile), middle-aged MPs (between the first and third quartile), and older MPs (more than the third quartile). The distribution plots for left- and right-wing speeches are presented in Figure \ref{fig:WingAge}.
We can observe that for most of the parliaments, the extreme-left speakers are predominantly middle-aged. The exception is Slovenia, where both younger and middle-aged speakers form the majority of extreme left-wing speakers. The extreme-right politicians in most parliaments are also predominantly middle-aged, except in Bulgaria, where the younger MPs form the majority of this group. 

\begin{figure}[!!htb]
  \centering
\begin{minipage}{0.45 \linewidth}
    \includegraphics[width=\linewidth]{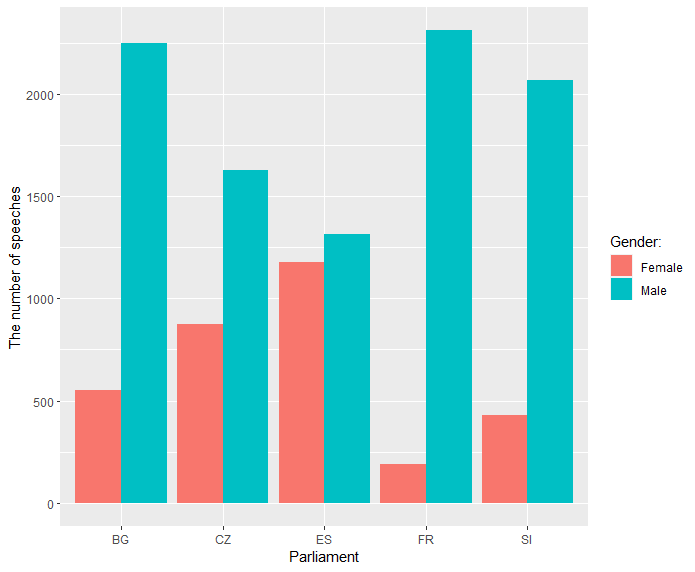} \\ \centerline{Extreme Left-wing speakers}
\end{minipage}
\begin{minipage}{0.45 \linewidth}
    \includegraphics[width=\linewidth]{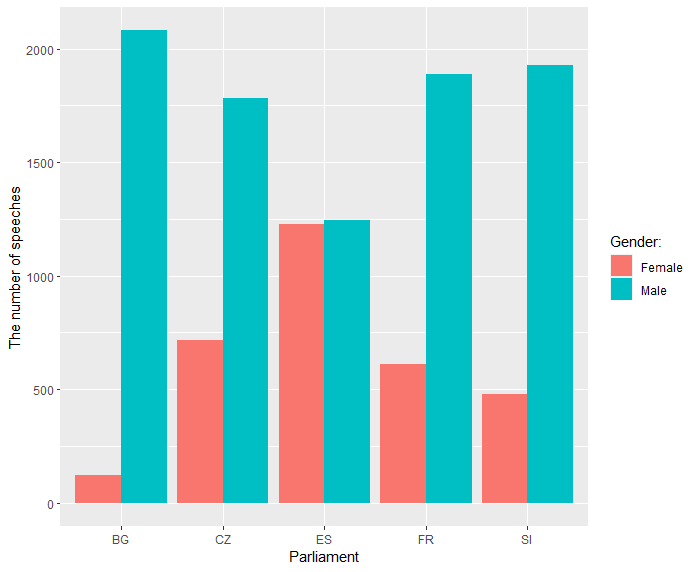} \\ \centerline{Extreme Right-wing speakers}
\end{minipage}
\caption{The distribution of the number of speeches relative to the MP's age group across the parliaments for left- and right-wing MPs. }
        \label{fig:WingGender}
\end{figure}

Similarly, we compared gender differences for both extreme-left and extreme-right positioned speakers in Figure \ref{fig:WingGender}. The difference in gender distributions are the most pronounced in Bulgaria and France. While for Bulgaria the number of female speakers drops going from extreme left to extreme-right, for France this number increases.

\section{Conclusions and Further Work}
\label{sec:conclusions}

We presented the mono- and cross-lingual methodology based on cross-domain transfer learning and Bayesian statistical testing for the analysis of parliamentary speeches. The proposed methodology and constructed models can be applied in a uniform way to the parliamentary speeches in the ParliaMint corpora collection (and other parliamentary datasets with similar information). Our methodology covers an analysis of sentiment and emotions, as well as prediction of metadata such as age, gender, and political orientation of the speakers. The source code of the developed methods and evaluation scenarios is publicly available\footnote{\url{ https://github.com/KristianMiok/Parliamentary-Discourse}}. We demonstrate the presented methodology on six national parliaments showing similarities and some surprising differences between them.

We discovered that in all countries except France, the age and gender of speakers are strong factors in the political discourse. Further, we found a big difference in the discourse between extreme left- and right-wing parties in all analyzed countries. Surprisingly, there is also a considerable difference between center-left and center-right parties in all countries except France. 
The sentiment analysis shows considerable differences between parliaments. The Czech, Spanish and United Kingdom parliaments express less negative than positive sentiment, the Bulgarian and French parliaments have a balanced distribution, and in the Slovenian parliament, the negative sentiment dominates. The sentiment is also significantly different for different age groups. The situation is different with the emotions, where positive emotions are strongly dominant in all countries except France and UK, where positive and negative emotions are almost balanced. 

There are many open avenues for further work. A larger analysis of all 16 parliaments in the ParliaMint collection would require a much larger research team who would be able to interpret the results but would produce a very interesting comparison between the parliaments. The proposed methodology could be extended with better training datasets for sentiment and emotions when they become available. We could analyze a broader spectrum of emotions, but currently, existing datasets are inadequate for our purpose due to differences in the covered domains.

\textbf{Funding Information} 

This work is based upon the collaboration in the COST Action CA18209 – NexusLinguarum “European network for Web-centred linguistic data science”, supported by COST (European Cooperation in Science and Technology). Marko Robnik-\v Sikonja received financial support from the Slovenian Research Agency through core research programme P6-0411 and projects J6-2581, J7-3159, and V5-2297. Encarnación Hidalgo Tenorio was financially supported by the European Social Fund, the Andalusian Government, and the University of Granada (Project References: A-HUM-250-UGR18 \& P18-FR-5020). Petya Osenova was partially supported by CLaDA-BG, the Bulgarian National Interdisciplinary Research e-Infrastructure for Resources and Technologies in favour of the Bulgarian Language and Cultural Heritage, Grant number DO01-377/18.12.2020.

\printbibliography

@book{maurer2001national,
  title={National parliaments on their ways to Europe. Losers or latecomers?},
  author={Maurer, Andreas and Wessels, Wolfgang},
  year={2001},
  publisher={Nomos Verlag}
}

@inproceedings{oberlander2018analysis,
  title={An analysis of annotated corpora for emotion classification in text},
  author={Oberl{\"a}nder, Laura Ana Maria and Klinger, Roman},
  booktitle={Proceedings of the 27th International Conference on Computational Linguistics},
  pages={2104--2119},
  year={2018}
}

@article{sailunaz2018emotion,
  title={Emotion detection from text and speech: a survey},
  author={Sailunaz, Kashfia and Dhaliwal, Manmeet and Rokne, Jon and Alhajj, Reda},
  journal={Social Network Analysis and Mining},
  volume={8},
  pages={1--26},
  year={2018},
  publisher={Springer}
}

@article{cortal2022natural,
  title={Natural Language Processing for Cognitive Analysis of Emotions},
  author={Cortal, Gustave and Finkel, Alain and Paroubek, Patrick and Ye, Lina},
  journal={arXiv preprint arXiv:2210.05296},
  year={2022}
}

@inproceedings{alm2005emotions,
  title={Emotions from text: machine learning for text-based emotion prediction},
  author={Alm, Cecilia Ovesdotter and Roth, Dan and Sproat, Richard},
  booktitle={Proceedings of human language technology conference and conference on empirical methods in natural language processing},
  pages={579--586},
  year={2005}
}

@article{han2007analysing,
  title={Analysing roll calls of the {European Parliament: A Bayesian} application},
  author={Han, Jeong-Hun},
  journal={European Union Politics},
  volume={8},
  number={4},
  pages={479--507},
  year={2007},
  xpublisher={Sage Publications Sage UK: London, England}
}

@article{hansen2009positions,
  title={The positions of {Irish} parliamentary parties 1937--2006},
  author={Hansen, Martin Ejnar},
  journal={Irish Political Studies},
  volume={24},
  number={1},
  pages={29--44},
  year={2009},
  publisher={Taylor \& Francis}
}

@article{miok2022ban,
  title={To {BAN} or not to {BAN}: {Bayesian} attention networks for reliable hate speech detection},
  author={Miok, Kristian and {\v{S}}krlj, Bla{\v{z}} and Zaharie, Daniela and Robnik-{\v{S}}ikonja, Marko},
  journal={Cognitive Computation},
  pages={1--19},
  year={2021},
  xpublisher={Springer}
}

@inproceedings{miok2019prediction,
  title={Prediction uncertainty estimation for hate speech classification},
  author={Miok, Kristian and Nguyen-Doan, Dong and {\v{S}}krlj, Bla{\v{z}} and Zaharie, Daniela and Robnik-{\v{S}}ikonja, Marko},
  booktitle={Statistical Language and Speech Processing: 7th International Conference, SLSP 2019, Ljubljana, Slovenia, October 14--16, 2019, Proceedings 7},
  pages={286--298},
  year={2019},
  organization={Springer}
}

@article{montalvo2019bayesian,
  title={Bayesian forecasting of electoral outcomes with new parties’ competition},
  author={Montalvo, Jos{\'e} G and Papaspiliopoulos, Omiros and Stumpf-F{\'e}tizon, Timoth{\'e}e},
  journal={European Journal of Political Economy},
  volume={59},
  pages={52--70},
  year={2019},
  publisher={Elsevier}
}

@inproceedings{ohman2020xed,
  title={{XED: A} Multilingual Dataset for Sentiment Analysis and Emotion Detection},
  author={{\"O}hman, Emily and P{\`a}mies, Marc and Kajava, Kaisla and Tiedemann, J{\"o}rg},
  booktitle={Proceedings of the 28th International Conference on Computational Linguistics},
  pages={6542--6552},
  year={2020}
}

@article{demszky2020goemotions,
  title={GoEmotions: A dataset of fine-grained emotions},
  author={Demszky, Dorottya and Movshovitz-Attias, Dana and Ko, Jeongwoo and Cowen, Alan and Nemade, Gaurav and Ravi, Sujith},
  journal={arXiv preprint arXiv:2005.00547},
  year={2020}
}

@inproceedings{saravia2018carer,
  title={Carer: Contextualized affect representations for emotion recognition},
  author={Saravia, Elvis and Liu, Hsien-Chi Toby and Huang, Yen-Hao and Wu, Junlin and Chen, Yi-Shin},
  booktitle={Proceedings of the 2018 conference on empirical methods in natural language processing},
  pages={3687--3697},
  year={2018}
}

@article{baraniak2021dataset,
  title={A dataset for sentiment analysis of entities in news headlines ({SEN})},
  author={Baraniak, Katarzyna and Sydow, Marcin},
  journal={Procedia Computer Science},
  volume={192},
  pages={3627--3636},
  year={2021},
  publisher={Elsevier}
}

@article{malo2014good,
  title={Good debt or bad debt: Detecting semantic orientations in economic texts},
  author={Malo, Pekka and Sinha, Ankur and Korhonen, Pekka and Wallenius, Jyrki and Takala, Pyry},
  journal={Journal of the Association for Information Science and Technology},
  volume={65},
  number={4},
  pages={782--796},
  year={2014},
  publisher={Wiley Online Library}
}

@article{buvcar2018annotated,
  title={Annotated news corpora and a lexicon for sentiment analysis in {Slovene}},
  author={Bu{\v{c}}ar, Jo{\v{z}}e and {\v{Z}}nidar{\v{s}}i{\v{c}}, Martin and Povh, Janez},
  journal={Language Resources and Evaluation},
  volume={52},
  number={3},
  pages={895--919},
  year={2018},
  publisher={Springer}
}

@inproceedings{glavavs2017unsupervised,
  title={Unsupervised Cross-Lingual Scaling of Political Texts},
  author={Glava{\v{s}}, Goran and Nanni, Federico and Ponzetto, Simone Paolo},
  booktitle={EACL 2017},
  pages={688},
  year={2017}
}

@article{iliev2019political,
  title={Political rhetoric through the lens of non-parametric statistics: are our legislators that different?},
  author={Iliev, Iliyan R and Huang, Xin and Gel, Yulia R},
  journal={Journal of the Royal Statistical Society: Series A (Statistics in Society)},
  volume={182},
  number={2},
  pages={583--604},
  year={2019},
  publisher={Wiley Online Library}
}

@article{hopkins2010method,
  title={A method of automated nonparametric content analysis for social science},
  author={Hopkins, Daniel J and King, Gary},
  journal={American Journal of Political Science},
  volume={54},
  number={1},
  pages={229--247},
  year={2010},
  publisher={Wiley Online Library}
}

@article{flaherty1987langue,
  title={Langue nationale/langue naturelle: {The} Politics of Linguistic Uniformity during the {French} Revolution},
  author={Flaherty, Peter},
  journal={Historical Reflections/R{\'e}flexions historiques},
  pages={311--328},
  year={1987},
  publisher={JSTOR}
}

@inproceedings{sakamoto2017cross,
  title={Cross-national measurement of polarization in political discourse: Analyzing floor debate in the US the Japanese legislatures},
  author={Sakamoto, Takuto and Takikawa, Hiroki},
  booktitle={2017 IEEE international conference on big data},
  pages={3104--3110},
  year={2017},
  xorganization={IEEE}
}

@incollection{honkela2014five,
  title={Five-dimensional sentiment analysis of corpora, documents and words},
  author={Honkela, Timo and Korhonen, Jaakko and Lagus, Krista and Saarinen, Esa},
  booktitle={Advances in self-organizing maps and learning vector quantization},
  pages={209--218},
  year={2014},
  publisher={Springer}
}

@article{rheault2016measuring,
  title={Measuring emotion in parliamentary debates with automated textual analysis},
  author={Rheault, Ludovic and Beelen, Kaspar and Cochrane, Christopher and Hirst, Graeme},
  journal={PloS ONE},
  volume={11},
  number={12},
  pages={e0168843},
  year={2016},
  xpublisher={Public Library of Science San Francisco, CA USA}
}

@inproceedings{dziecikatko2018application,
  title={Application of text analytics to analyze emotions in the speeches},
  author={Dzieciatko, Mariusz},
  booktitle={International Conference on Information Technologies in Biomedicine},
  pages={525--536},
  year={2018},
  organization={Springer}
}

@inproceedings{kowsari2020gender,
  title={Gender detection on social networks using ensemble deep learning},
  author={Kowsari, Kamran and Heidarysafa, Mojtaba and Odukoya, Tolu and Potter, Philip and Barnes, Laura E and Brown, Donald E},
  booktitle={Proceedings of the Future Technologies Conference},
  pages={346--358},
  year={2020},
  xorganization={Springer}
}

@inproceedings{menendez2020damegender,
  title={Damegender: Writing and Comparing Gender Detection Tools},
  author={Men{\'e}ndez, David Arroyo and Gonz{\'a}lez-Barahona, Jes{\'u}s M and Robles, Gregorio},
  booktitle={Proceedings of the Seminar Series on Advanced Techniques \& Tools for Software Evolution, SATToSE},
  year={2020}
}

@article{argamon2003gender,
  title={Gender, genre, and writing style in formal written texts},
  author={Argamon, Shlomo and Koppel, Moshe and Fine, Jonathan and Shimoni, Anat Rachel},
  journal={Text \& Talk},
  volume={23},
  number={3},
  pages={321--346},
  year={2003},
  publisher={De Gruyter Berlin, Germany}
}

@article{park2019gender,
  title={Gender classification using sentiment analysis and deep learning in a health web forum},
  author={Park, Sunghee and Woo, Jiyoung},
  journal={Applied Sciences},
  volume={9},
  number={6},
  pages={1249},
  year={2019},
  publisher={Multidisciplinary Digital Publishing Institute}
}

@article{de1997populism,
  title={Populism and democracy: political discourses and cultures in contemporary {Ecuador}},
  author={De la Torre, Carlos},
  journal={Latin American Perspectives},
  volume={24},
  number={3},
  pages={12--24},
  year={1997},
  xpublisher={Sage Publications Sage CA: Thousand Oaks, CA}
}

@article{elkink2021predicting,
  title={Predicting vote choice in the 2020 {Irish} general election},
  author={Elkink, Johan A and Farrell, David M},
  journal={Irish Political Studies},
  volume={36},
  number={4},
  pages={521--534},
  year={2021},
  xpublisher={Taylor \& Francis}
}

@inproceedings{abercrombie2020parlvote,
  title={{ParlVote: A} corpus for sentiment analysis of political debates},
  author={Abercrombie, Gavin and Batista-Navarro, Riza Theresa},
  booktitle={Proceedings of the 12th Language Resources and Evaluation Conference},
  pages={5073--5078},
  year={2020}
}

@inproceedings{rudkowskysupervised,
  title={Supervised Sentiment Analysis of Parliamentary Speeches and News Reports},
  author={Rudkowsky, Elena and Haselmayer, Martin and Wastian, Matthias and Jenny, Marcelo and Emrich, {\v{S}}tefan and Sedlmair, Michael},
  year={2017},
  booktitle={67th Annual Conference of the International Communication Association (ICA), Panel on Automatic Sentiment Analysis}
}

@inproceedings{van1985handbook,
  title={Handbook of discourse analysis},
  author={Van Dijk, Teun A},
  booktitle={Discourse and dialogue},
  year={1985},
  xorganization={Citeseer}
}

@article{trudgill1972sex,
  title={Sex, covert prestige and linguistic change in the urban British English of Norwich},
  author={Trudgill, Peter},
  journal={Language in society},
  volume={1},
  number={2},
  pages={179--195},
  year={1972},
  publisher={Cambridge University Press}
}

@article{salmela2017emotional,
  title={Emotional roots of right-wing political populism},
  author={Salmela, Mikko and Von Scheve, Christian},
  journal={Social Science Information},
  volume={56},
  number={4},
  pages={567--595},
  year={2017},
  publisher={SAGE Publications Sage UK: London, England}
}

@article{rodriguez1994youth,
  title={Youth and student slang in British and American English: an annotated bibliography},
  author={Rodriguez Gonzalez, Felix},
  journal={Revista alicantina de estudios ingleses},
  volume={7},
  pages={201-212},
  year={1994},
  xpublisher={Universidad de Alicante. Departamento de Filolog{\'\i}a Inglesa}
}

@book{murphy2010corpus,
  title={Corpus and sociolinguistics: Investigating age and gender in female talk},
  author={Murphy, Br{\'o}na},
  volume={38},
  year={2010},
  publisher={John Benjamins Publishing}
}

@article{milroy1985linguistic,
  title={Linguistic change, social network and speaker innovation1},
  author={Milroy, James and Milroy, Lesley},
  journal={Journal of linguistics},
  volume={21},
  number={2},
  pages={339--384},
  year={1985},
  publisher={Cambridge University Press}
}

@article{lakoff1973language,
  title={Language and woman's place},
  author={Lakoff, Robin},
  journal={Language in society},
  volume={2},
  number={1},
  pages={45--79},
  year={1973},
  xpublisher={Cambridge University Press}
}

@article{kerswill1996children,
  title={Children, adolescents, and language change},
  author={Kerswill, Paul},
  journal={Language variation and change},
  volume={8},
  number={2},
  pages={177--202},
  year={1996},
  publisher={Cambridge University Press}
}

@article{jay2013child,
  title={A child’s garden of curses: A gender, historical, and age-related evaluation of the taboo lexicon},
  author={Jay, Kristin L and Jay, Timothy B},
  journal={The American Journal of Psychology},
  volume={126},
  number={4},
  pages={459--475},
  year={2013},
  publisher={University of Illinois Press}
}

@article{tenorio2016genderlect,
  title={Genderlect},
  author={Hidalgo-Tenorio, E},
  journal={The Wiley Blackwell encyclopedia of gender and sexuality studies},
  year={2016},
  publisher={John Wiley \& Sons, Ltd}
}

@article{frizelle2018growth,
  title={Growth in syntactic complexity between four years and adulthood: evidence from a narrative task},
  author={Frizelle, Pauline and Thompson, Paul A and McDonald, David and Bishop, Dorothy VM},
  journal={Journal of Child Language},
  volume={45},
  number={5},
  pages={1174--1197},
  year={2018},
  publisher={Cambridge University Press}
}

@article{emara2017gender,
  title={Gender Identity Construction in {Facebook} Statuses of {Egyptian} Young Adults},
  author={Emara, Ingy},
  journal={Cairo Studies in English},
  volume={2017},
  number={1},
  pages={86--111},
  year={2017},
  publisher={Cairo University, Faculty of Arts, Department of English Language and Literature}
}

@article{de2012youth,
  title={“Youth Languages”: A Useful Invention?},
  author={de F{\'e}ral, Carole},
  journal={Langage et societe},
  number={3},
  volume = {141},
  pages={21--46},
  year={2012},
  xpublisher={Maison des sciences de l’homme}
}

@article{dahllof2012automatic,
  title={Automatic prediction of gender, political affiliation, and age in {Swedish} politicians from the wording of their speeches—{A} comparative study of classifiability},
  author={Dahll{\"o}f, Mats},
  journal={Literary and linguistic computing},
  volume={27},
  number={2},
  pages={139--153},
  year={2012},
  publisher={Oxford University Press}
}

@book{coates2015women,
  title={Women, men and language: A sociolinguistic account of gender differences in language},
  author={Coates, Jennifer},
  year={2015},
  publisher={Routledge}
}

@article{betz1993new,
  title={The new politics of resentment: {Radical} right-wing populist parties in {Western Europe}},
  author={Betz, Hans-George},
  journal={Comparative politics},
  pages={413--427},
  year={1993},
  publisher={JSTOR}
}

@book{barrett2020seven,
  title={Seven and a Half Lessons about the Brain},
  author={Barrett, Lisa Feldman},
  year={2020},
  publisher={Houghton Mifflin Harcourt Boston}
}

@article{alba2018emotion,
  title={Emotion and appraisal processes in language},
  author={Alba-Juez, Laura},
  journal={The Construction of Discourse as Verbal Interaction, Amsterdam, John Benjamins},
  pages={227--250},
  year={2018}
}

@article{ghafournia2015language,
  title={Language as a symbol of group membership},
  author={Ghafournia, Narjes},
  journal={Asian Social Science},
  volume={11},
  number={5},
  pages={19},
  year={2015},
  publisher={Citeseer}
}

@book{stenstrom2009youngspeak,
  title={Youngspeak in a multilingual perspective},
  author={Stenstr{\"o}m, Anna-Brita and J{\o}rgensen, Annette Myre},
  volume={184},
  year={2009},
  publisher={John Benjamins Publishing}
}

@book{labov1972sociolinguistic,
  title={Sociolinguistic patterns},
  author={Labov, William},
  xnumber={4},
  year={1972},
  publisher={University of Pennsylvania press}
}

@article{erjavec2022parlamint,
  title={The {ParlaMint} corpora of parliamentary proceedings},
  author={Erjavec, Tomaz and Ogrodniczuk, Maciej and Osenova, Petya and Ljubesic, Nikola and Simov, Kiril and Pancur, Andrej and Rudolf, Michal and Kopp, Matyas and Barkarson, Starkaur and Steingrmsson, Steinr and others},
  journal={Language resources and evaluation},
  pages={1--34},
  year={2022},
  publisher={Springer}
}

@inproceedings{devlin-etal-2019-bert,
    title = "{BERT}: Pre-training of Deep Bidirectional Transformers for Language Understanding",
    author = "Devlin, Jacob  and
      Chang, Ming-Wei  and
      Lee, Kenton  and
      Toutanova, Kristina",
    booktitle = "Proceedings of the 2019 Conference of the North {A}merican Chapter of the Association for Computational Linguistics: Human Language Technologies, Volume 1 (Long and Short Papers)",
    xmonth = jun,
    year = "2019",
    doi = "10.18653/v1/N19-1423",
    pages = "4171--4186"
}

@article{lehti2014style,
  title={Style in {French} politicians’ blogs: {Degree} of formality},
  author={Lehti, Lotta and Laippala, Veronika},
  journal={Language\@ internet},
  volume={11},
  number={1},
  year={2014}
}

@article{cameron201110,
  title={Aging, Age, and Sociolinguistics},
  author={Cameron, Richard},
  journal={The handbook of Hispanic sociolinguistics},
  pages={207},
  year={2011},
  publisher={John Wiley \& Sons}
}

@inproceedings{Vaswani2017,
  title={Attention is all you need},
  author={Vaswani, Ashish and Shazeer, Noam and Parmar, Niki and Uszkoreit, Jakob and Jones, Llion and Gomez, Aidan N and Kaiser, {\L}ukasz and Polosukhin, Illia},
  booktitle={Advances in neural information processing systems},
  pages={5998--6008},
  year={2017}
}

@article{RobnikSikonja2021, 
title={Cross-lingual transfer of sentiment classifiers},
volume={9}, 
url={https://journals.uni-lj.si/slovenscina2/article/view/9797}, 
DOI={10.4312/slo2.0.2021.1.1-25}, 
number={1}, 
journal={Slovenščina 2.0: empirical, applied and interdisciplinary research},
author={Robnik-Šikonja, Marko and Reba, Kristjan and Mozetič, Igor}, 
year={2021}, 
month={Jul.}, 
pages={1–25} 
}
%
%

\end{document}